\documentclass[]{fairmeta}

\usepackage{microtype}
\usepackage{graphicx}
\usepackage{subcaption}
\usepackage{booktabs} 
\usepackage{subcaption}
\usepackage{soul}

\usepackage{booktabs}
\usepackage{multirow}
\usepackage{graphicx}

\usepackage{hyperref}

\usepackage{booktabs, multirow, graphicx}
\usepackage[table]{xcolor}

\definecolor{dropGreen}{RGB}{198,239,206}
\definecolor{dropLightGreen}{RGB}{220,245,225}
\definecolor{dropLightRed}{RGB}{255,230,230}
\definecolor{dropRed}{RGB}{255,205,205}
\definecolor{dropDarkRed}{RGB}{255,175,175}
\definecolor{dropVDarkRed}{RGB}{255,140,140}

\usepackage{amsmath}
\usepackage{amssymb}
\usepackage{mathtools}
\usepackage{amsthm}

\usepackage[table]{xcolor}
\definecolor{LightGreen}{RGB}{220,245,220}
\definecolor{LightRed}{RGB}{245,220,220}

\theoremstyle{plain}

\theoremstyle{definition}

\theoremstyle{remark}

\usepackage{listings}

\usepackage{tikz}
\usetikzlibrary{positioning,calc,tikzmark}
\usepackage{xcolor}
\usepackage{pifont}   
\usepackage{enumitem} 
\usepackage{soul}
\newcommand{\cmark}{\textcolor{green!60!black}{\ding{51}}}
\newcommand{\xmark}{\textcolor{red!75!black}{\ding{55}}}

\usepackage[utf8]{inputenc} 
\usepackage[T1]{fontenc}    
\usepackage{hyperref}       
\usepackage{url}            
\usepackage{booktabs}       
\usepackage{amsfonts}       
\usepackage{nicefrac}       
\usepackage{microtype}      
\usepackage{xcolor}         
\usepackage{comment}
\usepackage[normalem]{ulem}

\title{Quantized Reasoning Models Think They Need to Think Longer, but They Do Not}

\author[1,]{Sanae Lotfi}
\author[1]{Polina Kirichenko}
\author[2]{Steven Li}
\author[2]{Zechun Liu}

\affiliation[1]{FAIR at Meta}
\affiliation[2]{Meta AI}

\abstract{
Post-training quantization (PTQ) is widely used to deploy large language models efficiently, but its effect on reasoning models is not well understood. Across math, coding, and science QA, we find that aggressive PTQ reduces accuracy while increasing chain-of-thought (CoT) length.
Surprisingly, we show that in up to $52\%$ of the quantized models' failures, models reach the right answer in intermediate reasoning steps but do not output it as a final answer.
To understand why quantization leads to this increase in overthinking errors, we measure the token-level KL divergence between quantized and full-precision output distributions.
Positions with high KL divergence correlate strongly with high next-token entropy, and at these positions quantized models disproportionately sample overthinking markers such as “wait'', “but'', and “alternatively''.
We show that simply introducing a training-free logit penalty on a curated set of overthinking markers can  reduce CoT length by 12--23\% while preserving or improving accuracy across 5 models (1.5B--32B parameters), 3 quantization methods, and 5 benchmarks, yielding a favorable Pareto frontier of accuracy against reasoning cost compared to penalizing other token sets.
Overthinking errors produced by quantized models are particularly reduced by up to 58\%.
}

\date{\today}
\correspondence{\email{sanaelotfi@meta.com}}
\metadata[Blogpost]{\url{https://sanaelotfi.github.io/blog/quantized-reasoning-models/}}

\begin{document}

\maketitle

\begin{figure*}[ht!]
    \centering
    \includegraphics[width=\linewidth]{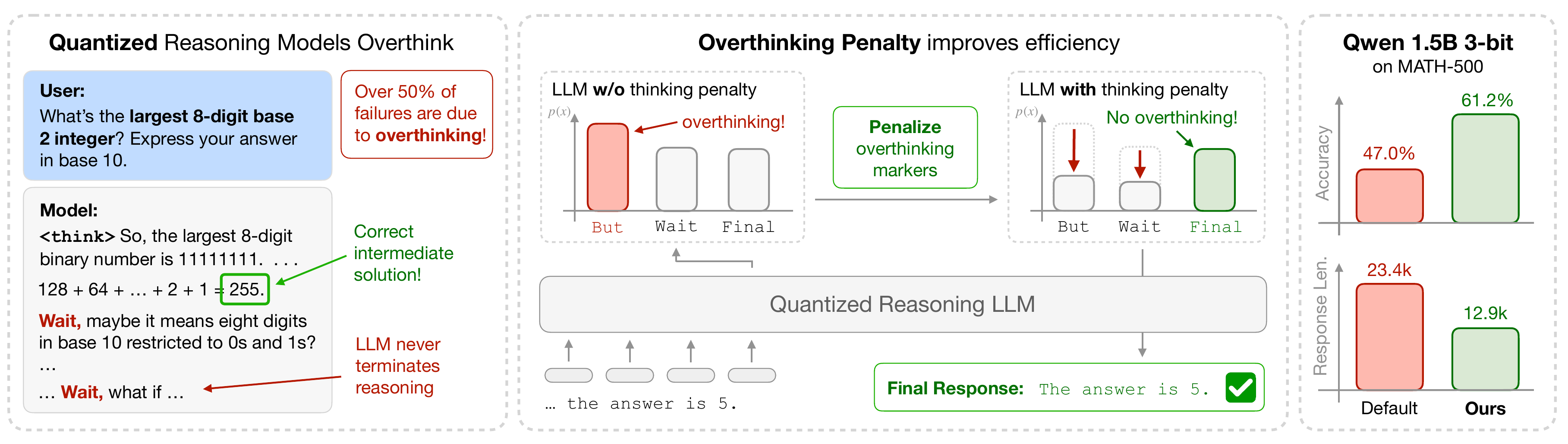}
    \caption{
    \textbf{Quantization exacerbates overthinking in reasoning models.}
    \textbf{(Left):} Even though quantized reasoning models reach an intermediate correct answer, they self-doubt, open too many new reasoning branches and never produce the final response.
    \textbf{(Middle):} We apply a logit penalty on thinking markers (i.e. ``but'' or ``wait'') which open new reasoning branches.
    \textbf{(Right):} We show that the overthinking penalty leads to significant improvements in efficiency (reasoning length) while maintaining or improving accuracy on average across models, datasets and quantization methods.
    }
    \label{fig:figure1}
\end{figure*}

\section{Introduction}

Post-training quantization (PTQ) is a widely adopted paradigm for efficient deployment of large language models \citep{lin2024awq, frantar2022gptq, xiao2023smoothquant, yao2022zeroquant}: it requires no gradient updates, no access to the training pipeline, and can be applied with limited calibration data.
When a quantized reasoning model fails to solve a problem, the natural assumption is that quantization has eroded the model's capability and the model simply can no longer find the correct answer.
In this work, we find that in up to 52\% of failures, the quantized model does produce the correct answer at some intermediate reasoning step, but then abandons it. Rather than committing to a correct conclusion, the model opens new reasoning branches, questions its own assumptions, and spirals into redundant deliberation that displaces the answer it had already found. Quantized reasoning models do not fail because they cannot think; they fail because they cannot \emph{stop} thinking.

This failure mode, which we refer to as \textbf{overthinking}, emerges as a consistent pattern in post-training quantized (PTQ) reasoning models.
While overthinking errors are also present in full-precision models, quantization substantially amplifies their frequency: overthinking accounts for 52\% of errors under AWQ 3-bit compared to 26\% in BF16, \textbf{a $7.3\times$ increase in the absolute number of overthinking errors}.
Motivated by this gap, we treat the full-precision model as a reference and analyze the token-level KL divergence between quantized and full-precision output distributions under identical generation prefixes to isolate the effects of quantization.

Two key findings emerge.
(1) The tokens with the highest average KL divergence between quantized and full-precision models contain \textit{overthinking markers} such as ``Wait,'' ``But,'' and ``Alternatively''. The tokens with the lowest KL divergence are mathematical and formatting tokens which encode the computational content of reasoning. The position-level KL divergence also correlates strongly with the next-token entropy of the full-precision model, confirming that quantization most affects positions where the model is already uncertain.
(2) Overthinking markers concentrate at high-entropy decoding positions as they appear 2--$4\times$ more frequently among the top-20 predictions at high-entropy steps than at low-entropy ones. Because these positions are inherently uncertain, quantization noise has a larger effect on the output distribution, further increasing the likelihood of sampling such tokens. This, in turn, triggers spurious reasoning branches that overwrite correct intermediate answers and delay termination.

To validate this diagnosis, we apply a simple training-free intervention: at each decoding step, a fixed logit penalty is applied to a curated token set $\mathcal{S}$. The penalty introduces zero computational overhead and only a single hyperparameter, yet is highly targeted. The selected tokens correspond to overthinking markers, which are strongly associated with high KL-divergence and high-entropy positions and lexically correspond to hesitation and backtracking. 
Across all configurations, the penalty consistently reduces CoT length by 12–23\% on average. Accuracy is preserved or often improved, reflecting a reduction in overthinking errors. In a few cases, accuracy slightly decreases, but CoT length is still significantly reduced, yielding a favorable shift in the efficiency–performance Pareto frontier.
Controlled ablations further highlight the specificity of the intervention: penalizing randomly selected tokens produces no consistent effect, while penalizing tokens with the lowest KL divergence leads to catastrophic degradation where the CoT length increases by up to 41\% and the accuracy drops by up to 9.5\%.
This specificity reflects how quantization degrades reasoning by amplifying path-forking tokens precisely at the high-uncertainty positions where they are most likely to induce repeated branching.

Our contributions are as follows: 
\begin{enumerate}
    \item We show that PTQ exacerbates overthinking in reasoning models. In up to 52\% of failures under aggressive quantization, the model reaches the correct answer in intermediate steps but fails to commit to it.
    \item We identify which tokens drive this failure through a KL divergence analysis between quantized and full-precision output distributions. Among the tokens with the highest divergence are hesitation and branching tokens that concentrate at high-entropy positions. The tokens with the lowest divergence are mathematical tokens that carry the computational content of reasoning.
    \item We validate this diagnosis with a  training-free logit penalty on a set of overthinking markers. The penalty consistently reduces CoT length by 12--23\% on average while preserving or improving accuracy. Controlled ablations confirm that penalizing overthinking markers yields the most favorable Pareto frontier of accuracy against reasoning cost.
\end{enumerate}


\section{Related Work}
\label{sec:related-work}

\textbf{Post-training quantization for LLMs.}
Post-training quantization can substantially reduce the model footprint while preserving perplexity and downstream performance.
Early work demonstrated robust 8-bit inference for large transformer models via outlier-aware mixed-precision strategies \citep{dettmers2022gpt3}. More recent PTQ methods target 4-bit and 3-bit weights for transformer blocks, including GPTQ \citep{frantar2022gptq}, which performs layerwise quantization using a second-order approximation of the loss landscape to correct for quantization errors, and AWQ \citep{lin2024awq}, which reduces quantization distortion by identifying and conserving important weight channels using activation statistics from calibration data. Other methods migrate the quantization difficulty from activations to weights such as SmoothQuant \citep{xiao2023smoothquant}, or are optimized to reduce quantization and dequantization overhead such as ZeroQuant \citep{yao2022zeroquant}.

Beyond these methods, several recent works improve low-bit PTQ via iterative error corrections. QuIP and QuIP\# propose principled low-bit quantization objectives to preserve accuracy at 2--4 bits \citep{chee2023quip,tseng2024quip}. Rotation-based approaches reduce outliers and improve quantization, including QuaRot~\citep{ashkboos2024quarot} and SpinQuant~\citep{liu2024spinquant}. FlatQuant enhances the flatness of weights and activations using optimal affine transformations including rotation and per-channel scaling~\citep{sun2024flatquant}. In our experiments, we focus on 3-bit and 4-bit weight-only quantization using AWQ and GPTQ, as well as 4-bit and 8-bit weight, activation, and KV-cache quantization using FlatQuant.

\textbf{Efficient reasoning.}
Overthinking in reasoning models is now a well-documented phenomenon. \citet{chen2024not} first quantified overthinking in o1-like LLMs, showing that models systematically over-allocate compute for simple problems, and proposed self-training strategies that reduce token usage by nearly 45\% without sacrificing accuracy. \citet{pipis2025wait} identify repetitive looping as a failure mode, particularly in smaller and distilled models. \citet{su2025between} show that the relationship between reasoning length and correctness is non-monotonic: longer traces do not always lead to better answers.
Mitigation methods range from fine-tuning models to self-brake before generating redundant steps \citep{zhao2025let} to dynamic early exit \citep{yang2025dynamic} and preferring the shortest correct chain among multiple samples \citep{hassid2025don}. We refer to \citet{sui2025stop} for a survey. Our approach differs in being focused on understanding how quantization worsens this overthinking behavior and how to mitigate it without any training, since PTQ itself is training-free.

\textbf{Effect of quantization on reasoning.}
Recent work suggests that the effect of quantization is metric- and task-dependent, and that compression can preserve aggregate performance while disproportionately degrading harder examples and capabilities \citep{hooker2020characterising, lotfi2024unlocking,hua2026uncertainty}. Several recent studies report that reasoning models degrade sharply under low-bit PTQ, exhibiting a drop in accuracy and an increase in chain-of-thought length \citep{liu2025quantization,mekala2025does,li2025quantization}. Our work is aligned with this emerging thread but differs in focus. Rather than only benchmarking accuracy under PTQ, we study how quantization changes reasoning traces at the token level, identifying an increase in overthinking errors and introducing a training-free intervention that addresses it.

\section{Experimental Setup}
\label{sec:setup}

\textbf{Quantization methods.} We evaluate post-training quantization under two settings. For weight-only quantization, we use GPTQ \citep{frantar2022gptq} and AWQ \citep{lin2024awq} at 3-bit and 4-bit precision with group size $g{=}128$. GPTQ performs layer-wise quantization using second-order information to minimize output reconstruction error. AWQ identifies salient weight channels using calibration activation statistics and applies per-channel rescaling to reduce quantization distortion. For end-to-end quantization of weights, activations, and the KV cache, we use FlatQuant \citep{sun2024flatquant} at W4A4KV4 and W8A8KV8 configurations. We use the notation W$b$A$c$KV$d$ to denote $b$-bit weights, $c$-bit activations, and $d$-bit KV cache. Additional details are provided in \Cref{app:exp-details}.

\textbf{Models.} We evaluate five reasoning-specialized LLMs spanning 1.5B to 32B parameters. From the DeepSeek-R1-Distill family, we use DeepSeek-R1-Distill-Qwen 1.5B, 7B, and 14B, as well as DeepSeek-R1-Distill-Llama 8B. These models are distilled from DeepSeek-R1 into Qwen-2.5 and Llama-3.1 architectures, respectively  \citep{Yang2024Qwen25TR}. We additionally evaluate QwQ-32B \citep{team2025qwq}, which acquires reasoning capabilities through reinforcement learning with verifiable rewards rather than distillation.

\textbf{Benchmarks.} We evaluate on five benchmarks spanning mathematics, coding, and science. For mathematics, we use GSM8K \citep{cobbe2021training} for shorter arithmetic chains, MATH-500 \citep{hendrycks2021measuring} for competition-style multi-step problems, and AIME-120 (which is composed of AIME-90\footnote{\url{https://huggingface.co/datasets/xiaoyuanliu/AIME90}} and AIME-2025 \citep{dekoninck2026matharena}) as a harder subset that tends to induce longer reasoning traces. For coding, we use LiveCodeBench \citep{jain2024livecodebench}. For science, we use GPQA-Diamond \citep{rein2024gpqa}.
All experiments use default decoding with $T{=}0.6$ and top-$p{=}0.95$, following the setting used by prior work on quantization of reasoning models \citep{liu2025quantization}.

\begin{figure}[t!]
\centering
\begin{subfigure}[t]{0.49\textwidth}
    \centering
    \includegraphics[width=\textwidth]{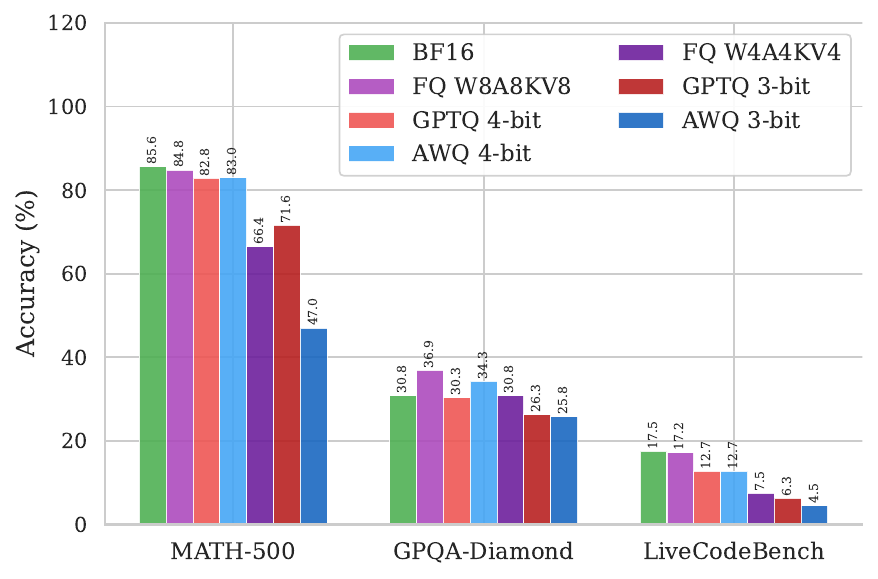}
    \caption{Accuracy drops under PTQ }
    \label{fig:ptq_acc}
\end{subfigure}
\hfill
\begin{subfigure}[t]{0.49\textwidth}
    \centering
    \includegraphics[width=\textwidth]{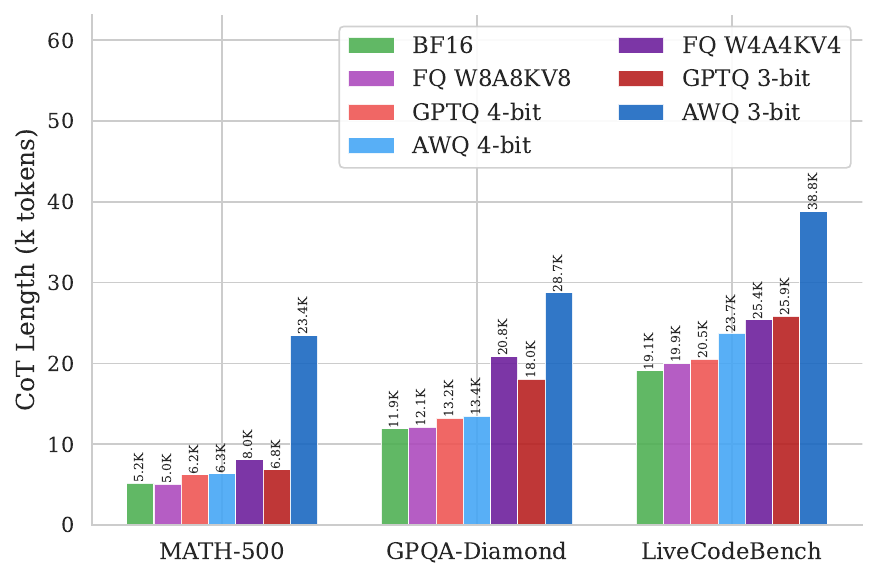}
    \caption{CoT length increases under PTQ }
    \label{fig:ptq_len}
\end{subfigure}
\caption{\textbf{Quantized reasoning models produce longer chains of thought while achieving lower accuracy.} We evaluate DeepSeek-R1-Distill-Qwen-1.5B in full precision (BF16) and after quantization on MATH-500, GPQA-Diamond, and LiveCodeBench. \textbf{(a)} Accuracy decreases as quantization becomes more aggressive, with 3-bit AWQ dropping MATH-500 accuracy from 85.6\% to 47.0\%. \textbf{(b)} CoT length increases significantly under aggressive quantization: 3-bit AWQ increases MATH-500 CoT from 5.2K to 23.4K tokens, a $4.5\times$ increase. Results for all five models are in \Cref{app:add-results}.
}
\label{fig:ptq_acc_len}
\end{figure}

\section{Quantization Exacerbates Overthinking in Reasoning Models}
\label{sec:quant-overthink}

\subsection{Accuracy drops while CoT length increases under PTQ}
\label{sec:acc-len}

First, we study how PTQ affects both accuracy and CoT length on reasoning tasks.
\Cref{fig:ptq_acc_len} shows the effect of progressively more aggressive quantization on DeepSeek-R1-Distill-Qwen-1.5B across three benchmarks.
Mild quantization (FlatQuant W8A8KV8 and 4-bit weight-only AWQ and GPTQ) largely preserves accuracy, with CoT lengths remaining close to the BF16 baseline.
As precision decreases further, both accuracy and reasoning efficiency degrade together.
On MATH-500, 3-bit AWQ reduces accuracy from 85.6\% to 47.0\% while increasing the average CoT from 5.2K to 23.4K tokens, a $4.5\times$ increase. GPTQ at 3-bit is more robust on the same dataset, retaining 71.6\% accuracy with 6.8K tokens, but still produces 32\% longer reasoning traces than BF16.
FlatQuant W4A4KV4, which quantizes weights, activations, and the KV cache, falls between the two: accuracy drops to 66.4\% and CoT grows to 8.0K tokens.

We observe the same pattern across four additional models (Qwen-7B, Qwen-14B, Llama-8B, QwQ-32B) and report full results in \Cref{app:add-results}. Aggressive quantization (3-bit weight-only and W4A4KV4) consistently leads to longer reasoning traces, with the effect most pronounced for smaller models and harder benchmarks.
On LiveCodeBench, 3-bit AWQ increases Qwen-1.5B CoT from 19.1K to 38.8K tokens while accuracy drops from 17.5\% to 4.5\%. Larger models are more resilient: Qwen-14B under 3-bit AWQ loses only 4.0\% average accuracy, and its CoT increase is modest by comparison.

We compute the Spearman correlation coefficient $\rho$ between accuracy degradation and CoT length increase across all 28 model-quantization pairs and find $\rho = -0.73$.
In other words, quantized models with the largest accuracy loss tend to also generate the longest reasoning traces. Given this strong correlation, we hypothesize that the additional reasoning induced by quantization is not just a co-occurring effect of lower accuracy, but that the longer chains of thought actively contribute to the performance degradation.
We investigate this hypothesis next.

\begin{figure*}[t!]
  \centering

  \begin{subfigure}[t]{0.49\textwidth}
    \centering
    \includegraphics[width=\linewidth]{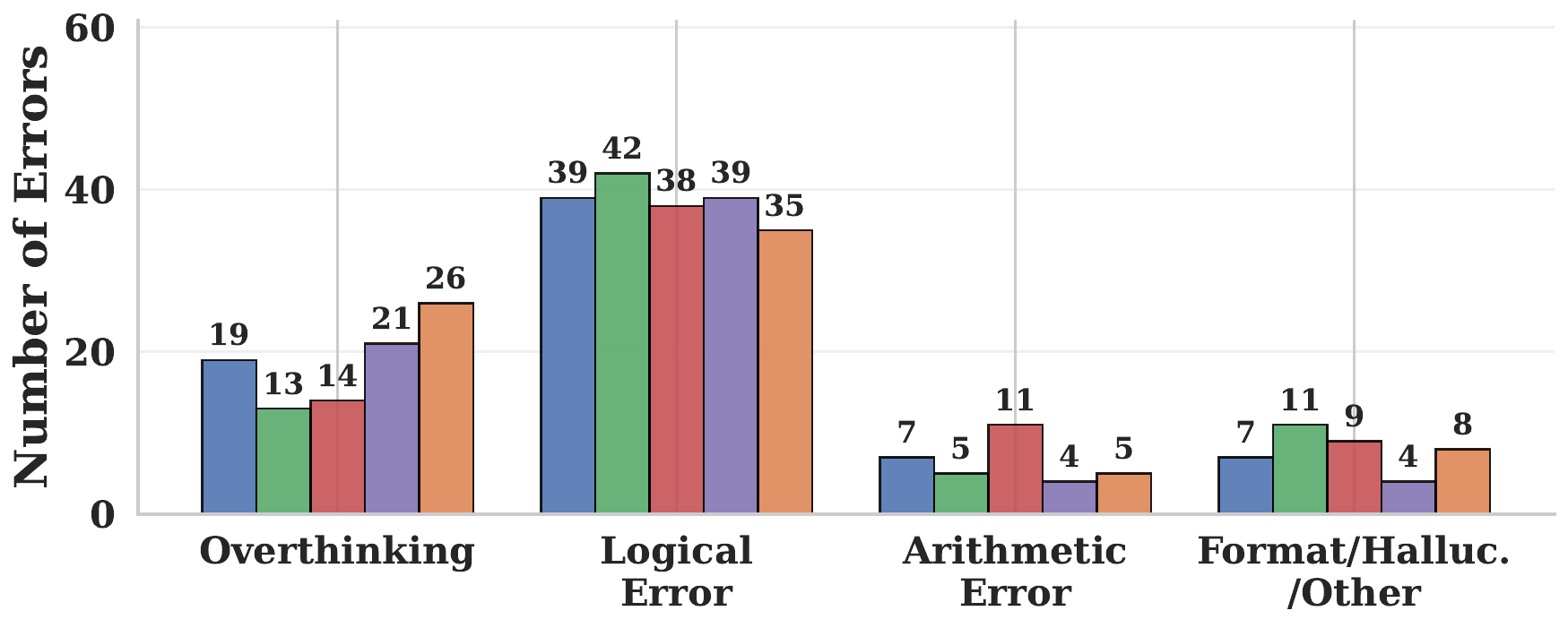}
    \caption{\small BF16}
  \end{subfigure}\hfill
  \begin{subfigure}[t]{0.49\textwidth}
    \centering
    \includegraphics[width=\linewidth]{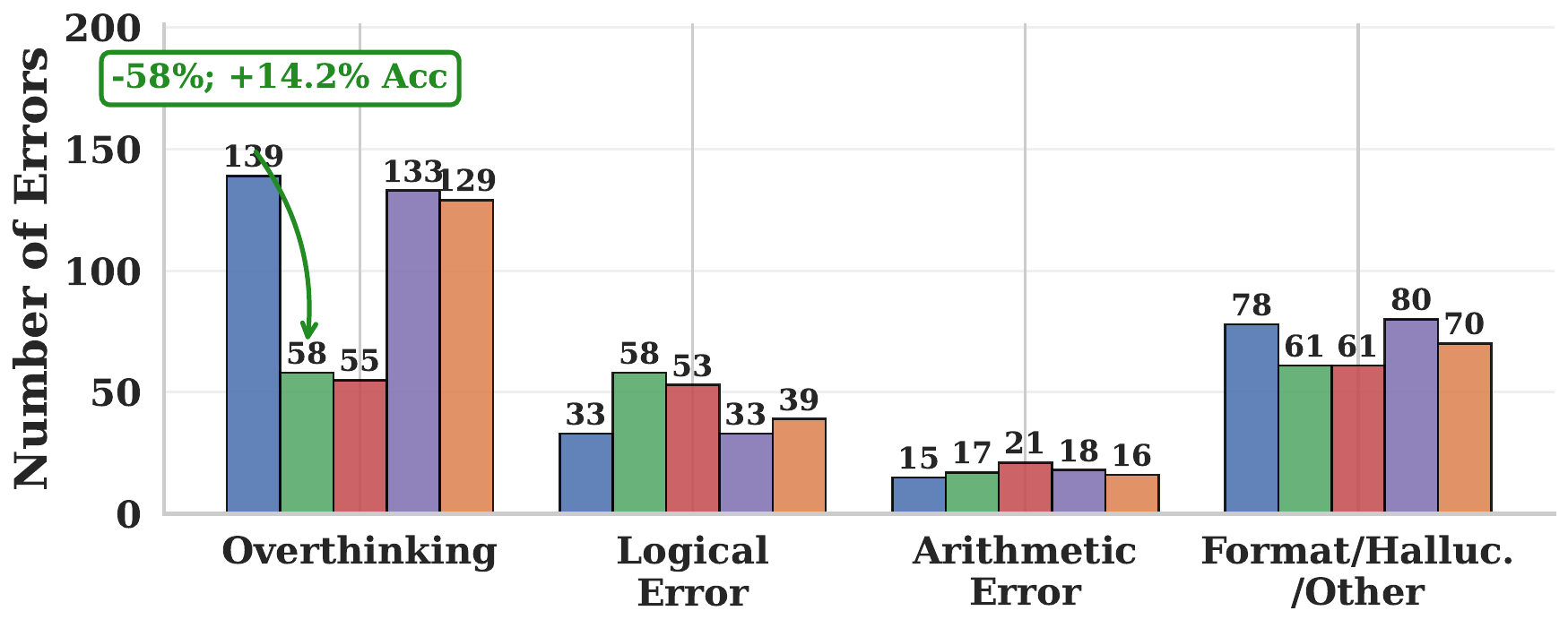}
    \caption{\small 3-bit AWQ}
  \end{subfigure}

  \begin{subfigure}[t]{0.49\textwidth}
    \centering
    \includegraphics[width=\linewidth]{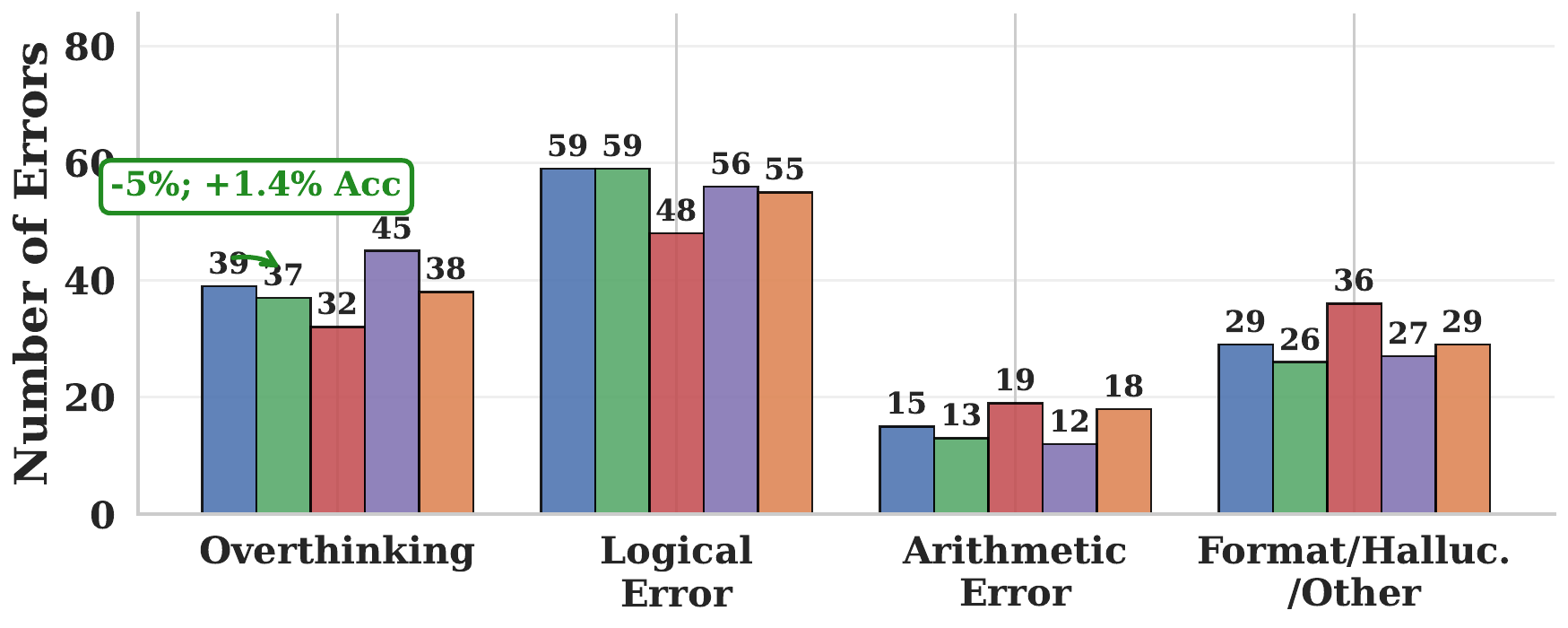}
    \caption{\small 3-bit GPTQ}
  \end{subfigure}\hfill
  \begin{subfigure}[t]{0.49\textwidth}
    \centering
    \includegraphics[width=\linewidth]{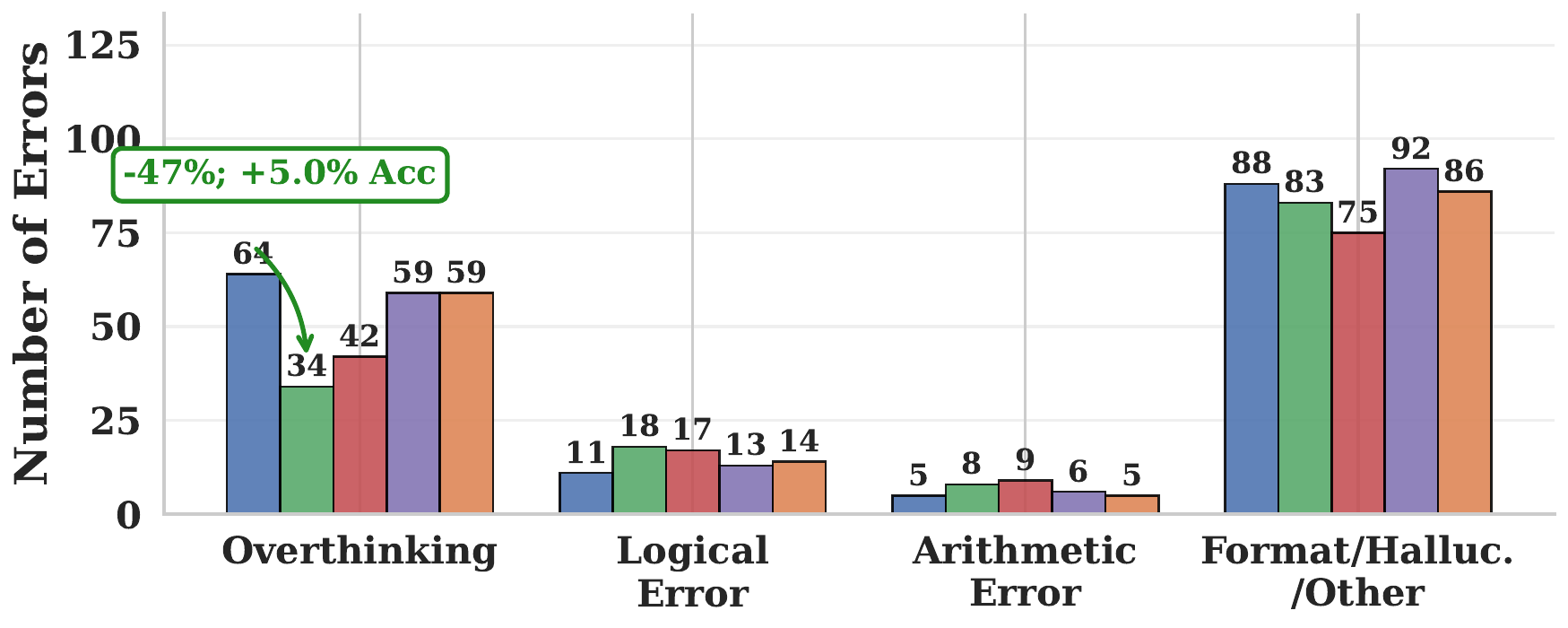}
    \caption{\small FlatQuant W4A4KV4}
  \end{subfigure}

  \includegraphics[width=0.89\textwidth]{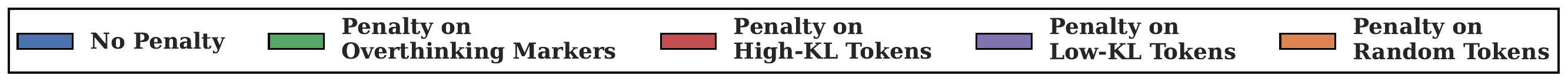}

  \caption{\textbf{Quantization increases overthinking errors, and penalizing overthinking markers reduces them.} We categorize errors from DeepSeek-R1-Distill-Qwen-1.5B on MATH-500 into four types. Each panel shows the error count under no penalty and four penalty configurations. Overthinking errors increase under quantization. Penalizing overthinking markers and high-KL tokens consistently reduces overthinking errors by up to 58\% on AWQ 3-bit while improving accuracy by up to 14.2\%. Penalizing random or low-KL tokens does not reduce overthinking errors and can even increase such errors.
  }
  \label{fig:error_breakdown}
\end{figure*}

\subsection{Quantization increases overthinking errors}
\label{sec:error-analysis}

To understand what kind of errors quantization leads to, we categorize incorrect answers from DeepSeek-R1-Distill-Qwen-1.5B on MATH-500. We first annotate a subset of errors manually, then use GPT-5 \citep{singh2025openai} as an LLM judge to scale the categorization after tuning the prompt to achieve over 95\% agreement with human annotation. The prompt is provided in \Cref{app:prompt-error-categorization}.

We assign each failure to exactly one of the following categories: (i) \textbf{Overthinking.} The model reaches the correct solution at some point in its chain of thought but does not commit to it as the final answer. Instead, the model opens new reasoning paths, excessively questions its own assumptions, or reverses a correct conclusion. A representative example is shown in \Cref{fig:qual_overthinking}; (ii) \textbf{Logical error.} The model follows an incorrect plan or misunderstands the problem from early on, such that the reasoning trajectory is wrong before any correct solution is reached; (iii) \textbf{Arithmetic error.} The overall approach and plan are correct, but the model makes concrete computational or calculation mistakes that lead to a wrong final answer; and (iv) \textbf{Formatting, hallucination, and other errors.} The model produces an off-format answer, introduces hallucinated constraints not present in the problem, or exhibits other failures not covered by the categories above.

\Cref{fig:error_breakdown} shows the error breakdown for BF16 and three aggressive quantization methods. In BF16, the model makes 72 errors out of 500 questions, with only 19 attributed to overthinking (26\%). Quantization increases overthinking errors across all three methods. Under 3-bit AWQ, overthinking errors rise from 19 to 139, a $7.3\times$ increase, making overthinking the dominant failure mode at 52\% of all 265 errors. GPTQ at 3-bit doubles the overthinking count from 19 to 39. FlatQuant more than triples it from 19 to 64. In all three cases, quantization results in an increase in errors where the model reaches the correct answer but fails to commit to it. We show similar results for GSM8K in \Cref{fig:error_gsm8k}.

This increase in overthinking errors under quantization supports our hypothesis from \Cref{sec:acc-len}. The longer reasoning traces produced by quantized models contain excessive hesitation and self-doubt that often turn correct intermediate answers into incorrect final answers.

\begin{figure*}[t!]
\centering
\begin{subfigure}[t]{0.325\textwidth}
    \centering
    \includegraphics[width=\textwidth]{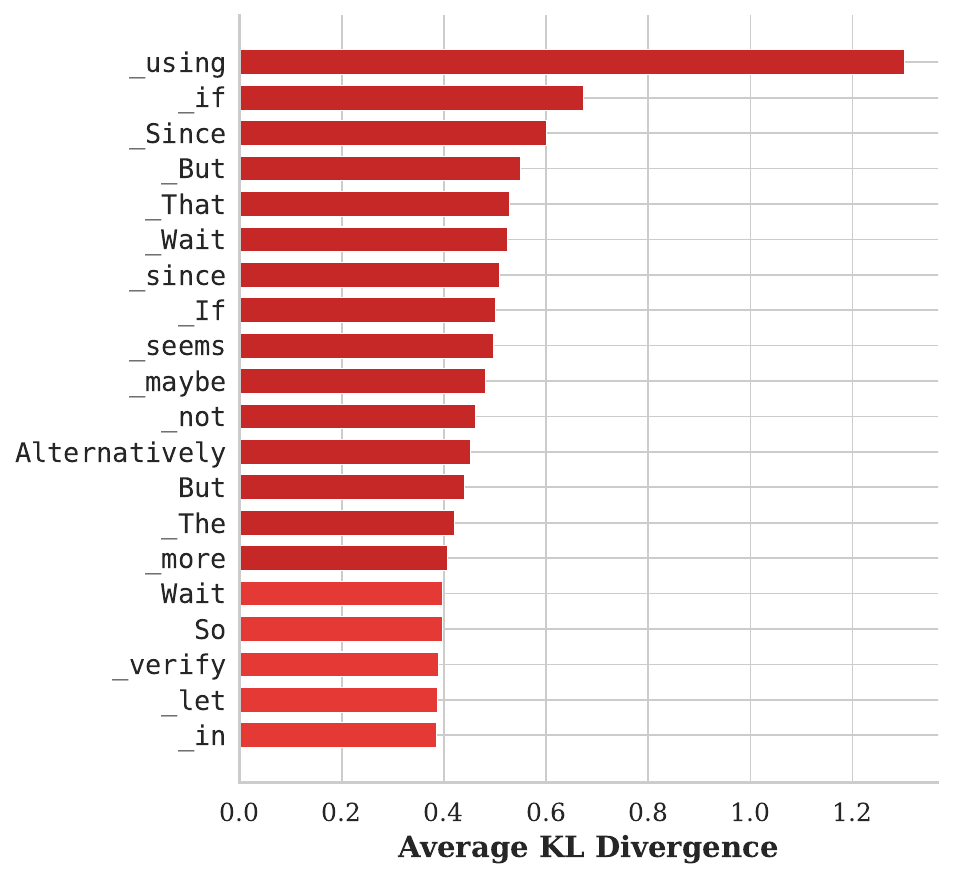}
    \caption{Highest KL tokens}
    \label{fig:kl_high}
\end{subfigure}
\hfill
\begin{subfigure}[t]{0.325\textwidth}
    \centering
    \includegraphics[width=\textwidth]{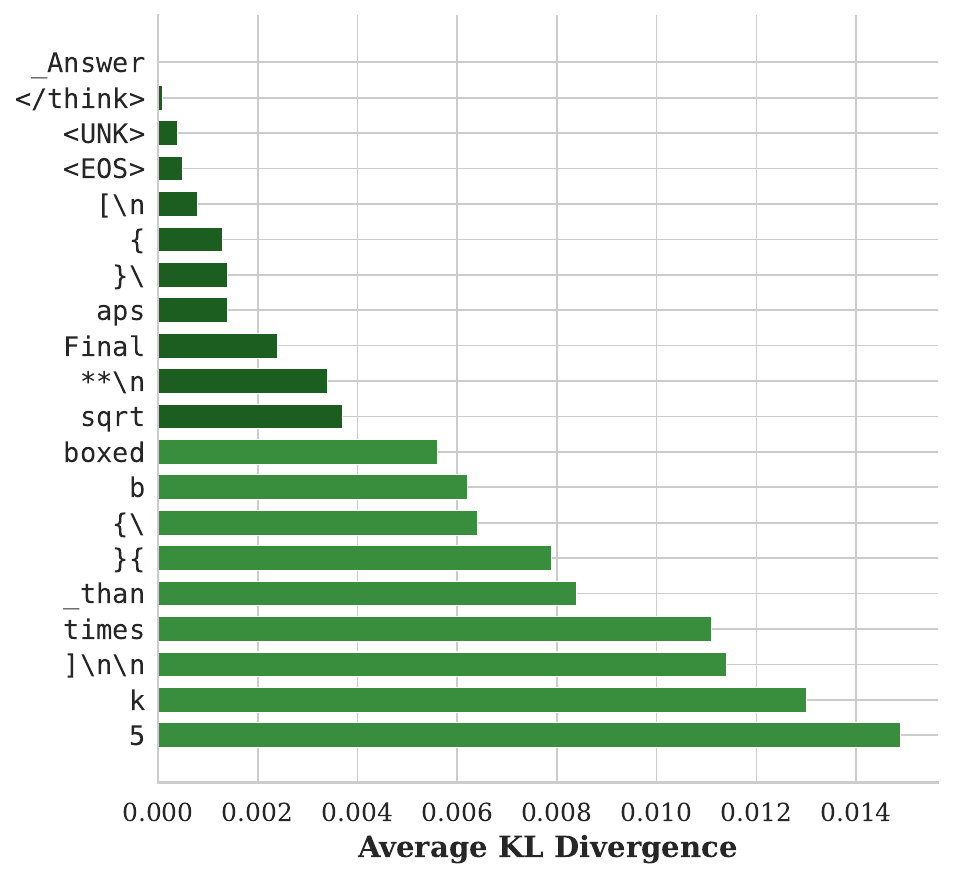}
    \caption{Lowest KL tokens}
    \label{fig:kl_low}
\end{subfigure}
\hfill
\begin{subfigure}[t]{0.325\textwidth}
    \centering
    \includegraphics[width=\textwidth]{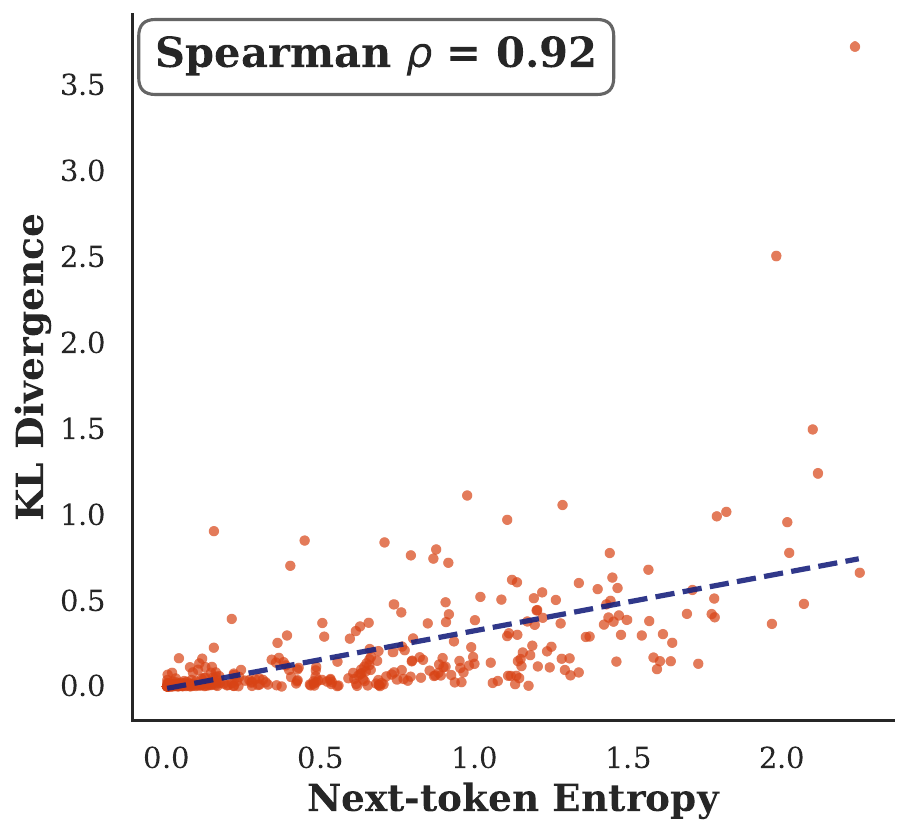}
    \caption{KL vs next-token entropy}
    \label{fig:kl_entropy}
\end{subfigure}
\caption{\textbf{Quantization disproportionately affects thinking and hesitation tokens at high-entropy positions.} We compute the average KL divergence between the BF16 and 3-bit AWQ output distributions for each token appearing at least 50 times on MATH-500. (a) The 20 highest-KL tokens are thinking and branching tokens. (b) The 20 lowest-KL tokens are mathematical and formatting tokens. Tokens prefixed with ``\_'' include a leading space in the tokenizer. (c) Position-level KL divergence correlates strongly with the BF16 model's next-token entropy (Spearman $\rho = 0.92$), confirming that quantization most affects positions where the model is already uncertain.}
\label{fig:kl_tokens}
\end{figure*}

\subsection{Quantized models diverge from non-quantized ones in branching positions}
\label{sec:kl-analysis}

To understand \textit{how} quantization increases overthinking errors, we measure the token-level KL divergence between the output distributions of the BF16 and 3-bit AWQ models.
We treat the full-precision model as a reference and run both models on the same MATH-500 prompts under identical generation prefixes to isolate the effects of quantization.
At each decoding position $t$, we compute the KL divergence $D_{\mathrm{KL}}(p_t \| q_t) = \sum_{v \in \mathcal{V}} p_t(v) \log \frac{p_t(v)}{q_t(v)}$, where $p_t$ and $q_t$ are the next-token distributions of the full-precision and quantized models, respectively.
We then associate each KL value with the token that the quantized model sampled at that position and compute, for each unique token in the vocabulary, its average KL divergence across all positions where it was sampled. We filter to tokens appearing at least 50 times across all generations.

\Cref{fig:kl_tokens} shows the 20 tokens with the highest and lowest associated average KL divergence. The tokens where quantized and full-precision models diverge the most are thinking and branching tokens such as ``Wait'', ``But'', ``Alternatively'', ``if'', ``maybe'', ``verify'', ``think''.
Many of these tokens open new reasoning paths and express the model's hesitation.
The tokens where the two models agree most closely are mathematical and formatting tokens.
Quantization barely changes how the model samples these execution tokens.
We provide word cloud visualizations of the full set of 50 highest-KL and 50 lowest-KL tokens in \Cref{app:wordclouds}.

We also compute the density of high-KL and low-KL tokens relative to total CoT length. In BF16, low-KL math tokens appear at a higher rate than high-KL hesitation tokens, with a ratio of 0.57. Under 3-bit AWQ, this ratio flips to 1.15: high-KL tokens outnumber low-KL tokens, meaning the quantized model produces more thinking and hesitation tokens per unit of reasoning than mathematical content. This shift in token composition explains how quantization leads to less efficient reasoning.

\section{Penalizing Overthinking Markers}
\label{sec:penalty}

The analysis in \Cref{sec:kl-analysis} shows that quantization disproportionately affects tokens associated with hesitation and redirection, while leaving mathematical tokens largely unchanged. We now test a simple inference-time intervention that reduces the probability of these tokens during decoding.

\subsection{Overthinking penalty}
\label{sec:method}

We curate a set $\mathcal{S}$ of 50 \emph{overthinking markers}: tokens that express hesitation, self-doubt, or reasoning redirection. These include tokens such as ``Wait'', ``But'', ``Alternatively'', ``perhaps'', ``maybe'', ``however'', ``reconsider'', ``backtrack'', and ``wrong''. The full list is provided in \Cref{app:thinking-tokens}.

We construct $\mathcal{S}$ by selecting tokens that signal the model is opening a new reasoning branch or questioning a conclusion it has already reached. Several recent works have identified similar tokens as indicators of redundant reasoning in language models \citep{zhao2025let, fan2025missing}.
We deliberately exclude tokens from the high-KL list analyzed in \Cref{sec:kl-analysis} that serve roles unrelated to overthinking, such as ``so'' which would in fact help the model conclude its reasoning.

At each decoding step $t$, we modify the logits $z_t(v)$ for each token $v \in \mathcal{S}$ in our overthinking markers list by subtracting a fixed penalty $\lambda > 0$:
\begin{equation}
z'_t(v) = \begin{cases}
z_t(v) - \lambda & \text{if } v \in \mathcal{S}, \\
z_t(v) & \text{otherwise}.
\end{cases}
\label{eq:penalty}
\end{equation}

The penalty does not require additional forward passes and adds no computational overhead. The only hyperparameter is $\lambda$, which controls the strength of the overthinking penalty.

\begin{figure*}[t]
\centering
\begin{subfigure}[t]{0.325\textwidth}
    \centering
    \includegraphics[width=\textwidth]{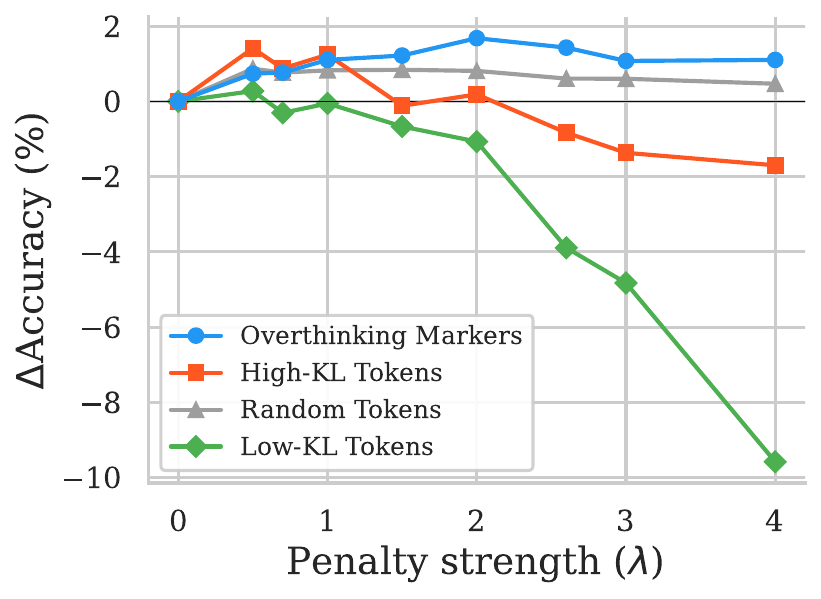}
    \caption{Accuracy vs $\lambda$}
    \label{fig:acc_vs_lambda}
\end{subfigure}
\hfill
\begin{subfigure}[t]{0.325\textwidth}
    \centering
    \includegraphics[width=\textwidth]{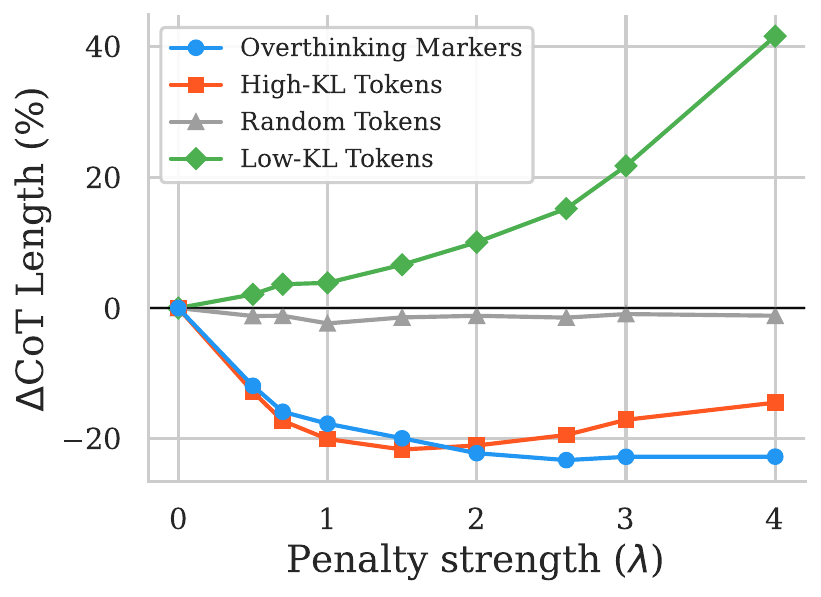}
    \caption{CoT length vs $\lambda$}
    \label{fig:len_vs_lambda}
\end{subfigure}
\hfill
\begin{subfigure}[t]{0.325\textwidth}
    \centering
    \includegraphics[width=\textwidth]{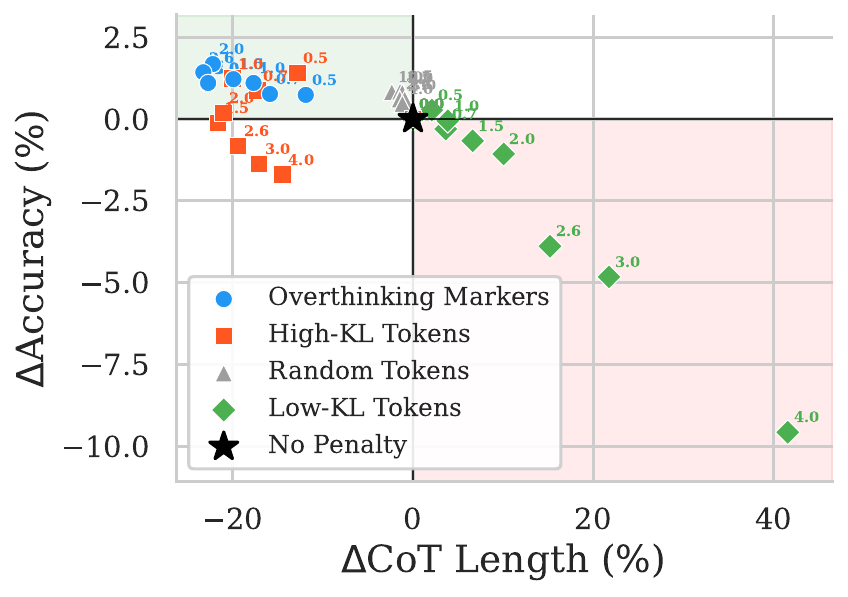}
    \caption{Pareto frontier}
    \label{fig:pareto}
\end{subfigure}
\caption{\textbf{Penalizing overthinking markers consistently reduces CoT length while preserving accuracy.} We compare four token lists (50 tokens each) for the logit penalties on Qwen-1.5B, averaged across six quantization configurations and five benchmarks. (a) Penalizing overthinking markers maintains or improves accuracy across all $\lambda$ values, while penalizing low-KL tokens degrades accuracy.  (b) Overthinking markers consistently reduce CoT length by 12\% to 23\%. Low-KL tokens increase CoT by up to 41\%. Penalizing random tokens has negligible effect. (c) Overthinking markers occupy the upper-left region (shorter CoT, better accuracy) across all $\lambda$ values, yielding the best efficiency-accuracy tradeoff. Black star marks the no-penalty baseline, and penalizing random tokens remains close to this baseline. Note that the numbers on the markers reflect the penalty strength.}
\label{fig:ablation}
\end{figure*}

\begin{table}[t!]
\centering
\caption{\textbf{Penalizing overthinking markers reduces CoT length significantly while maintaining or improving accuracy on average under extreme quantization.} For each configuration, the top row (Base) shows accuracy (\%) and CoT length (thousands of tokens) without penalty, and the bottom row (+Pen) shows results with the best $\lambda$; we report the standard deviation w.r.t $\lambda$ in the $\Delta$ columns. Our overthinking penalty consistently reduces CoT length across all configurations, yielding a better Pareto frontier of accuracy against reasoning cost than the baseline. In several cases accuracy also improves.
\textbf{Per-benchmark cells are colored when the penalty changes accuracy or CoT length by $\geq$2\%.}
We show the most aggressive quantization settings; and report the rest in \Cref{app:add-results}.}
\label{tab:main_results}
\resizebox{\textwidth}{!}{%
\setlength{\tabcolsep}{4pt}
\renewcommand{\arraystretch}{1.15}
\small
\begin{tabular}{cl@{\hskip 4pt}l ccccc c cc}
\toprule
 & &  & \textbf{AIME} & \textbf{MATH} & \textbf{GSM8K} & \textbf{GPQA} & \textbf{LCB} & \textbf{Avg.} & \textbf{$\Delta$Acc} & \textbf{$\Delta$Len\%} \\
\cmidrule(lr){10-10} \cmidrule(lr){11-11}
 & &  & \multicolumn{5}{c}{\footnotesize\textit{Acc (\%)\;/\;Len (k)}} & \footnotesize\textit{Acc\;/\;Len} & \multicolumn{2}{c}{\footnotesize\textit{mean$_{\pm\text{std}}$}} \\
\midrule
\multirow{8}{*}{\rotatebox{90}{Qwen-1.5B}} & \multirow{2}{*}{\textbf{BF16}} & Base & \textbf{24.4}\,/\,\textbf{17.5} & \textbf{85.6}\,/\,\textbf{5.2} & \textbf{84.2}\,/\,\textbf{2.7} & \textbf{30.8}\,/\,\textbf{11.9} & \textbf{17.5}\,/\,\textbf{19.1} & \textbf{48.5}\,/\,\textbf{11.2} & & \\
 & & +Pen & 25.6\,/\,\colorbox{green!17}{14.8} & 85.8\,/\,\colorbox{green!15}{4.5} & 85.1\,/\,\colorbox{green!9}{2.6} & \colorbox{green!35}{39.4}\,/\,\colorbox{green!14}{10.5} & 17.5\,/\,\colorbox{green!10}{18.3} & 50.7\,/\,10.1 & \cellcolor{green!24}$+2.1_{\scriptscriptstyle\pm 1.6}$ & \cellcolor{green!19}$-9.2_{\scriptscriptstyle\pm 5.7}$ \\
\cline{2-11}
 & \multirow{2}{*}{GPTQ W3} & Base & 8.9\,/\,20.5 & 71.6\,/\,6.8 & 76.1\,/\,4.0 & 26.3\,/\,18.0 & 6.3\,/\,25.9 & 37.8\,/\,15.0 & & \\
 & & +Pen & \colorbox{green!16}{11.1}\,/\,\colorbox{green!22}{15.6} & 73.0\,/\,\colorbox{green!16}{5.8} & 77.5\,/\,\colorbox{green!26}{2.8} & \colorbox{green!17}{28.8}\,/\,\colorbox{green!26}{12.5} & 8.2\,/\,\colorbox{green!16}{22.2} & 39.7\,/\,11.8 & \cellcolor{green!23}$+1.9_{\scriptscriptstyle\pm 1.1}$ & \cellcolor{green!35}$-22.6_{\scriptscriptstyle\pm 6.2}$ \\
\cline{2-11}
 & \multirow{2}{*}{AWQ W3} & Base & 4.4\,/\,33.0 & 47.0\,/\,23.4 & 63.5\,/\,16.3 & 25.8\,/\,28.7 & 4.5\,/\,38.8 & 29.0\,/\,28.0 & & \\
 & & +Pen & \colorbox{green!23}{8.9}\,/\,\colorbox{green!14}{29.3} & \colorbox{green!40}{61.2}\,/\,\colorbox{green!35}{12.9} & \colorbox{green!28}{69.5}\,/\,\colorbox{green!35}{7.6} & \colorbox{green!23}{30.3}\,/\,\colorbox{green!20}{22.7} & 6.3\,/\,\colorbox{green!13}{35.3} & 35.2\,/\,21.6 & \cellcolor{green!45}$+6.2_{\scriptscriptstyle\pm 1.7}$ & \cellcolor{green!41}$-28.0_{\scriptscriptstyle\pm 7.5}$ \\
\cline{2-11}
 & \multirow{2}{*}{FQ W4A4KV4} & Base & 7.8\,/\,17.5 & 66.4\,/\,8.0 & 78.5\,/\,3.4 & 30.8\,/\,20.8 & 7.5\,/\,25.4 & 38.2\,/\,15.0 & & \\
 & & +Pen & \colorbox{green!23}{12.2}\,/\,\colorbox{green!10}{16.8} & \colorbox{green!25}{71.4}\,/\,\colorbox{green!21}{6.2} & 78.9\,/\,\colorbox{green!31}{2.1} & 31.3\,/\,\colorbox{green!17}{17.6} & 8.6\,/\,\colorbox{green!11}{23.8} & 40.5\,/\,13.3 & \cellcolor{green!25}$+2.3_{\scriptscriptstyle\pm 1.3}$ & \cellcolor{green!29}$-17.6_{\scriptscriptstyle\pm 4.9}$ \\
\midrule
\multirow{8}{*}{\rotatebox{90}{Qwen-7B}} & \multirow{2}{*}{\textbf{BF16}} & Base & \textbf{45.6}\,/\,\textbf{12.7} & \textbf{93.2}\,/\,\textbf{3.9} & \textbf{91.1}\,/\,\textbf{1.9} & \textbf{50.0}\,/\,\textbf{9.9} & \textbf{37.7}\,/\,\textbf{16.5} & \textbf{63.5}\,/\,\textbf{9.0} & & \\
 & & +Pen & \colorbox{green!29}{52.2}\,/\,\colorbox{green!15}{11.2} & 94.2\,/\,\colorbox{green!14}{3.5} & 91.6\,/\,\colorbox{green!15}{1.6} & \colorbox{green!29}{56.6}\,/\,\colorbox{green!10}{9.4} & 38.8\,/\,\colorbox{green!13}{14.9} & 66.7\,/\,8.1 & \cellcolor{green!32}$+3.2_{\scriptscriptstyle\pm 1.9}$ & \cellcolor{green!20}$-10.0_{\scriptscriptstyle\pm 2.7}$ \\
\cline{2-11}
 & \multirow{2}{*}{GPTQ W3} & Base & 30.0\,/\,17.0 & 91.0\,/\,4.8 & 90.1\,/\,2.5 & 41.9\,/\,11.4 & 27.2\,/\,19.1 & 56.0\,/\,11.0 & & \\
 & & +Pen & \colorbox{green!23}{34.4}\,/\,\colorbox{green!15}{14.8} & 91.2\,/\,\colorbox{green!15}{4.2} & 90.8\,/\,\colorbox{green!26}{1.8} & \colorbox{green!29}{48.5}\,/\,\colorbox{green!14}{10.3} & 26.5\,/\,19.2 & 58.3\,/\,10.1 & \cellcolor{green!25}$+2.2_{\scriptscriptstyle\pm 1.5}$ & \cellcolor{green!23}$-13.1_{\scriptscriptstyle\pm 3.0}$ \\
\cline{2-11}
 & \multirow{2}{*}{AWQ W3} & Base & 30.0\,/\,16.1 & 91.2\,/\,4.8 & 90.4\,/\,2.1 & 48.0\,/\,10.7 & 26.9\,/\,21.3 & 57.3\,/\,11.0 & & \\
 & & +Pen & 28.9\,/\,\colorbox{green!13}{14.6} & 90.2\,/\,\colorbox{green!18}{4.0} & 90.5\,/\,\colorbox{green!17}{1.8} & 49.0\,/\,\colorbox{green!11}{10.1} & 28.7\,/\,\colorbox{green!14}{18.9} & 57.5\,/\,9.9 & $+0.2_{\scriptscriptstyle\pm 1.4}$ & \cellcolor{green!22}$-12.0_{\scriptscriptstyle\pm 3.9}$ \\
\cline{2-11}
 & \multirow{2}{*}{FQ W4A4KV4} & Base & 23.3\,/\,20.4 & 83.6\,/\,5.4 & 87.7\,/\,2.1 & 50.0\,/\,18.0 & 12.3\,/\,23.9 & 51.4\,/\,14.0 & & \\
 & & +Pen & 22.2\,/\,\colorbox{green!10}{19.7} & \colorbox{green!18}{86.4}\,/\,\colorbox{green!15}{4.7} & 88.5\,/\,\colorbox{green!9}{2.1} & 51.5\,/\,\colorbox{green!18}{14.9} & 14.2\,/\,\colorbox{green!9}{23.2} & 52.6\,/\,12.9 & \cellcolor{green!18}$+1.2_{\scriptscriptstyle\pm 2.4}$ & \cellcolor{green!17}$-7.7_{\scriptscriptstyle\pm 4.8}$ \\
\midrule
\multirow{8}{*}{\rotatebox{90}{Qwen-14B}} & \multirow{2}{*}{\textbf{BF16}} & Base & \textbf{57.8}\,/\,\textbf{13.1} & \textbf{94.6}\,/\,\textbf{3.7} & \textbf{94.1}\,/\,\textbf{1.7} & \textbf{57.6}\,/\,\textbf{9.5} & \textbf{49.6}\,/\,\textbf{14.4} & \textbf{70.7}\,/\,\textbf{8.5} & & \\
 & & +Pen & \colorbox{green!26}{63.3}\,/\,\colorbox{green!19}{10.7} & 95.6\,/\,\colorbox{green!11}{3.5} & 94.5\,/\,\colorbox{green!10}{1.7} & \colorbox{green!25}{62.6}\,/\,\colorbox{green!12}{8.8} & \colorbox{green!24}{54.5}\,/\,\colorbox{green!12}{13.4} & 74.1\,/\,7.6 & \cellcolor{green!33}$+3.4_{\scriptscriptstyle\pm 1.7}$ & \cellcolor{green!18}$-8.5_{\scriptscriptstyle\pm 2.3}$ \\
\cline{2-11}
 & \multirow{2}{*}{GPTQ W3} & Base & 44.4\,/\,12.4 & 94.4\,/\,3.6 & 92.8\,/\,1.6 & 53.5\,/\,9.1 & 42.2\,/\,14.6 & 65.5\,/\,8.3 & & \\
 & & +Pen & \colorbox{green!20}{47.8}\,/\,\colorbox{green!10}{11.9} & 93.8\,/\,\colorbox{green!11}{3.4} & 92.5\,/\,\colorbox{green!13}{1.5} & \colorbox{green!17}{56.1}\,/\,\colorbox{green!11}{8.7} & \colorbox{green!25}{47.4}\,/\,14.6 & 67.5\,/\,8.0 & \cellcolor{green!24}$+2.0_{\scriptscriptstyle\pm 1.7}$ & \cellcolor{green!13}$-4.8_{\scriptscriptstyle\pm 2.3}$ \\
\cline{2-11}
 & \multirow{2}{*}{AWQ W3} & Base & 48.9\,/\,14.3 & 93.0\,/\,4.3 & 93.0\,/\,2.1 & 56.6\,/\,10.5 & 42.2\,/\,17.5 & 66.7\,/\,9.7 & & \\
 & & +Pen & 47.8\,/\,14.1 & 94.2\,/\,\colorbox{green!13}{3.9} & 92.9\,/\,\colorbox{green!18}{1.7} & 56.1\,/\,\colorbox{green!9}{10.2} & \colorbox{green!21}{45.9}\,/\,\colorbox{green!12}{16.2} & 67.4\,/\,9.2 & \cellcolor{green!14}$+0.6_{\scriptscriptstyle\pm 1.7}$ & \cellcolor{green!16}$-7.5_{\scriptscriptstyle\pm 3.5}$ \\
\cline{2-11}
 & \multirow{2}{*}{FQ W4A4KV4} & Base & 52.2\,/\,12.7 & 95.0\,/\,3.7 & 93.5\,/\,1.8 & 63.6\,/\,8.8 & 49.3\,/\,15.2 & 70.7\,/\,8.4 & & \\
 & & +Pen & 53.3\,/\,12.6 & 94.6\,/\,\colorbox{green!9}{3.6} & 93.9\,/\,\colorbox{green!15}{1.6} & \colorbox{red!16}{61.6}\,/\,8.8 & 49.3\,/\,\colorbox{green!11}{14.3} & 70.5\,/\,8.2 & $-0.2_{\scriptscriptstyle\pm 1.3}$ & \cellcolor{green!13}$-4.5_{\scriptscriptstyle\pm 3.5}$ \\
\midrule
\multirow{8}{*}{\rotatebox{90}{Llama-8B}} & \multirow{2}{*}{\textbf{BF16}} & Base & \textbf{44.4}\,/\,\textbf{14.5} & \textbf{89.0}\,/\,\textbf{4.6} & \textbf{89.0}\,/\,\textbf{2.2} & \textbf{49.0}\,/\,\textbf{10.2} & \textbf{35.1}\,/\,\textbf{16.2} & \textbf{61.3}\,/\,\textbf{9.6} & & \\
 & & +Pen & \colorbox{red!16}{42.2}\,/\,\colorbox{green!11}{13.8} & \colorbox{green!16}{91.0}\,/\,\colorbox{green!13}{4.2} & 88.6\,/\,\colorbox{green!12}{2.0} & 50.5\,/\,10.1 & \colorbox{green!19}{38.4}\,/\,\colorbox{green!11}{15.3} & 62.1\,/\,9.1 & \cellcolor{green!15}$+0.8_{\scriptscriptstyle\pm 2.0}$ & \cellcolor{green!14}$-5.6_{\scriptscriptstyle\pm 2.7}$ \\
\cline{2-11}
 & \multirow{2}{*}{GPTQ W3} & Base & 26.7\,/\,18.2 & 82.2\,/\,6.2 & 85.4\,/\,2.4 & 29.8\,/\,12.8 & 24.6\,/\,24.1 & 49.7\,/\,12.7 & & \\
 & & +Pen & \colorbox{green!16}{28.9}\,/\,\colorbox{green!14}{16.2} & 83.4\,/\,\colorbox{green!13}{5.6} & 85.8\,/\,\colorbox{green!15}{2.0} & \colorbox{green!40}{39.9}\,/\,12.7 & \colorbox{green!16}{26.9}\,/\,\colorbox{green!10}{23.0} & 53.0\,/\,11.9 & \cellcolor{green!32}$+3.2_{\scriptscriptstyle\pm 1.9}$ & \cellcolor{green!17}$-7.9_{\scriptscriptstyle\pm 3.9}$ \\
\cline{2-11}
 & \multirow{2}{*}{AWQ W3} & Base & 12.2\,/\,14.8 & 80.4\,/\,5.4 & 82.6\,/\,2.4 & 35.9\,/\,9.6 & 23.1\,/\,18.1 & 46.8\,/\,10.1 & & \\
 & & +Pen & \colorbox{green!23}{16.7}\,/\,\colorbox{green!12}{13.6} & \colorbox{red!18}{77.6}\,/\,\colorbox{green!15}{4.7} & 83.7\,/\,\colorbox{green!18}{2.0} & 36.9\,/\,\colorbox{green!14}{8.6} & 23.1\,/\,\colorbox{green!15}{15.9} & 47.6\,/\,9.0 & \cellcolor{green!15}$+0.7_{\scriptscriptstyle\pm 1.5}$ & \cellcolor{green!22}$-12.2_{\scriptscriptstyle\pm 2.7}$ \\
\cline{2-11}
 & \multirow{2}{*}{FQ W4A4KV4} & Base & 22.2\,/\,15.4 & 86.4\,/\,5.0 & 87.6\,/\,2.4 & 40.9\,/\,10.2 & 35.1\,/\,17.8 & 54.4\,/\,10.2 & & \\
 & & +Pen & 22.2\,/\,\colorbox{green!10}{14.7} & \colorbox{red!16}{84.2}\,/\,4.9 & 86.1\,/\,\colorbox{green!12}{2.2} & \colorbox{green!32}{48.5}\,/\,\colorbox{green!10}{9.8} & \colorbox{red!20}{31.7}\,/\,\colorbox{green!9}{17.3} & 54.5\,/\,9.8 & $+0.1_{\scriptscriptstyle\pm 1.7}$ & \cellcolor{green!12}$-4.1_{\scriptscriptstyle\pm 3.0}$ \\
\midrule
\multirow{6}{*}{\rotatebox{90}{QwQ-32B}} & \multirow{2}{*}{\textbf{BF16}} & Base & \textbf{76.7}\,/\,\textbf{15.3} & \textbf{97.2}\,/\,\textbf{4.5} & \textbf{95.7}\,/\,\textbf{1.9} & \textbf{63.1}\,/\,\textbf{10.1} & \textbf{62.3}\,/\,\textbf{17.9} & \textbf{79.0}\,/\,\textbf{9.9} & & \\
 & & +Pen & \colorbox{green!23}{81.1}\,/\,\colorbox{green!19}{12.3} & 98.0\,/\,\colorbox{green!14}{4.0} & 96.0\,/\,\colorbox{green!12}{1.8} & \colorbox{green!20}{66.7}\,/\,\colorbox{green!11}{9.5} & 61.9\,/\,\colorbox{green!10}{17.3} & 80.7\,/\,9.0 & \cellcolor{green!22}$+1.7_{\scriptscriptstyle\pm 1.2}$ & \cellcolor{green!19}$-9.6_{\scriptscriptstyle\pm 2.6}$ \\
\cline{2-11}
 & \multirow{2}{*}{GPTQ W3} & Base & 64.4\,/\,16.1 & 96.8\,/\,4.9 & 95.4\,/\,2.0 & 56.1\,/\,9.3 & 50.4\,/\,19.1 & 72.6\,/\,10.3 & & \\
 & & +Pen & 65.6\,/\,\colorbox{green!12}{15.0} & 96.4\,/\,\colorbox{green!14}{4.3} & 95.9\,/\,\colorbox{green!12}{1.8} & \colorbox{green!22}{60.1}\,/\,\colorbox{red!9}{9.5} & 51.9\,/\,\colorbox{green!11}{18.0} & 74.0\,/\,9.7 & \cellcolor{green!19}$+1.3_{\scriptscriptstyle\pm 1.4}$ & \cellcolor{green!15}$-5.9_{\scriptscriptstyle\pm 3.9}$ \\
\cline{2-11}
 & \multirow{2}{*}{AWQ W3} & Base & 60.0\,/\,16.7 & 96.0\,/\,4.6 & 95.9\,/\,1.8 & 61.1\,/\,10.1 & 51.5\,/\,19.5 & 72.9\,/\,10.5 & & \\
 & & +Pen & \colorbox{green!16}{62.2}\,/\,\colorbox{green!15}{14.5} & 96.6\,/\,\colorbox{green!13}{4.2} & 95.5\,/\,\colorbox{green!12}{1.7} & 60.6\,/\,\colorbox{green!16}{8.7} & 53.4\,/\,\colorbox{green!10}{18.7} & 73.7\,/\,9.5 & \cellcolor{green!15}$+0.7_{\scriptscriptstyle\pm 1.2}$ & \cellcolor{green!19}$-9.6_{\scriptscriptstyle\pm 2.9}$ \\
\bottomrule
\end{tabular}%
}
\end{table}

\subsection{Penalizing overthinking markers reduces CoT while preserving accuracy on average}
\label{sec:ablation}

We compare the effect of penalizing four different token lists of equal size (50 tokens each): our overthinking markers, the tokens with the highest KL divergence between BF16 and AWQ 3-bit (high-KL tokens), the tokens with the lowest KL divergence (low-KL tokens), and 50 randomly selected tokens. We apply the logit penalty as defined in \Cref{eq:penalty} to each list when evaluating Qwen-1.5B across six quantization configurations and five benchmarks, sweeping $\lambda$ from 0.5 to 4.0.

\Cref{fig:ablation} shows the results averaged across all configurations and benchmarks. Penalizing overthinking markers consistently reduces CoT length by 12\% to 23\% across all $\lambda$ values while maintaining or slightly improving accuracy. Penalizing random tokens has negligible effect on both accuracy and CoT length. Penalizing high-KL tokens reduces CoT at low $\lambda$ but degrades accuracy at higher values, since this list includes tokens that could actually help terminate the reasoning. Penalizing low-KL tokens is catastrophic: CoT increases by up to 41\% and accuracy drops by up to 9.5\%, confirming that these tokens are not the ones causing post-quantization degradation.

We further verify the directionality of the penalty by running a symmetric sweep with negative $\lambda$ values that boost rather than suppress the target tokens as shown in \Cref{fig:neg_sweep}. Boosting overthinking markers increases CoT length by up to 445\% and drops accuracy by up to 34\%. High-KL tokens show a comparable effect, while random tokens remain neutral. These results confirm that overthinking markers and high-KL tokens are the tokens that control the reasoning length in both directions.

We show in \Cref{fig:pareto} that overthinking markers occupy the upper-left region of the Pareto frontier \textit{for all $\lambda$ values}, achieving shorter CoT without sacrificing accuracy. No other token list achieves this: random tokens cluster near the no-penalty origin, high-KL tokens drift toward worse accuracy at strong penalties, and low-KL tokens move into the lower-right region where both accuracy and efficiency degrade.
We also note that the overthinking penalty is robust to the choice of $\lambda$: every $\lambda$ value tested reduces CoT length by at least 12\% on average across configurations on Qwen-1.5B while preserving performance as in \Cref{fig:acc_vs_lambda,fig:len_vs_lambda}.
In \Cref{tab:main_results}, we report results with the best $\lambda$ per configuration alongside the standard deviation across all $\lambda$ values.

\subsection{Overthinking penalty improves efficiency across models}
\label{sec:penalty-results}

\Cref{tab:main_results} shows the effect of the overthinking penalty across five models and the most aggressive quantization settings. The penalty reduces CoT length by 4.1\% to 28.0\% on average across benchmarks.
In most cases, accuracy is preserved or slightly improved.
Notably, there are a few cases where the accuracy deteriorates but the CoT length improves.
On AWQ 3-bit with Qwen-1.5B, accuracy improves by 6.2\% while CoT decreases by 28.0\%. 
Larger models show smaller but consistent CoT reductions, improving the efficiency of quantized models and partially reversing the overthinking behavior exacerbated by quantization. 
We report full results for all quantization methods in \Cref{app:add-results}.

\Cref{fig:error_breakdown} shows that the penalty on overthinking markers reduces the number of cases where the model reaches the correct answer but fails to commit. On AWQ 3-bit, overthinking errors drop from 139 to 58, a 58\% reduction, while total errors decrease from 265 to 194. On FlatQuant W4A4KV4, overthinking errors drop from 64 to 34 (47\% reduction) with total errors decreasing from 168 to 143. Penalizing random tokens or low-KL tokens can actually lead to an increase in overthinking errors, demonstrating the effectiveness of the penalty over our curated list of overthinking markers.
We observe the same pattern on GSM8K, where penalizing overthinking markers reduces overthinking errors across the three extreme quantization configurations from 344 to 237, a 31\% reduction as shown in \Cref{app:error-gsm8k}.

The overthinking penalty also reduces CoT length for BF16 models, with average reductions between 5.6\% and 10.0\% depending on the model. Quantization does not introduce overthinking as a new failure mode but rather exacerbates a tendency already present in full-precision models, as reflected by both the CoT increase after quantization and the increase of overthinking errors in particular.

\section{Why Does Quantization Exacerbate Overthinking?}
\label{sec:discussion}

In the previous sections we demonstrated that quantization increases overthinking errors and CoT length, and that penalizing overthinking markers addresses both. We now discuss and empirically verify a hypothesis for \textit{why} quantization worsens overthinking in reasoning models.

\citet{yayla2026mcel} show that the robustness of a quantized neural network to parameter perturbations is governed by the output-layer logit margin: the gap $m = z_{(1)} - z_{(2)}$ between the top-1 and top-2 logits. When this margin is large, a perturbation to the model's weights is unlikely to flip the argmax; when the margin is small, even a moderate perturbation can change the selected output. While their analysis targets classification networks under deterministic bit-flip errors, the same structural argument applies to the autoregressive setting under stochastic quantization noise \citep{widrow1996statistical, defossez2021differentiable}: locally uncertain positions can be affected more aggressively by quantization noise.

We verify this intuition in the autoregressive setting. We show in \Cref{fig:kl_entropy} that the correlation coefficient between the KL divergence computed on MATH-500 under identical generation prefixes and the next-token entropy of the full-precision model is $\rho = 0.92$ across all positions, confirming that positions where the model is most uncertain (small logit gap, high entropy) are where the quantized and full-precision models diverge the most.

The connection to overthinking follows from a second empirical observation: overthinking markers are 2 to $4\times$ more likely to appear among the top-20 most probable tokens at high-entropy positions compared to low-entropy positions as shown in \Cref{app-fig:top-tokens}. At these uncertain positions, quantization noise can shift the sampled token toward an overthinking marker such as ``Wait'' or ``But'', causing the model to open a new reasoning branch instead of continuing its current path. We note that this argument provides intuition for one specific failure mode, not a complete causal explanation of all types of degradations caused by PTQ.

\section{Conclusion, Limitations, and Future Work}
\label{sec:conclusion}

In this work, we studied the effect of post-training quantization on reasoning models and found that aggressive quantization exacerbates overthinking, where the model reaches correct intermediate answers but fails to commit to them, leading to longer chains of thought and lower accuracy. Through a token-level KL divergence analysis between quantized and full-precision models, we identified that hesitation and branching tokens are disproportionately affected by quantization while mathematical tokens remain stable. A simple training-free logit penalty on 50 curated overthinking markers consistently reduces CoT length by 12--23\% leading to a better accuracy-reasoning cost Pareto frontier across five models, three quantization methods, and five benchmarks.

Overthinking is not unique to quantized models. BF16 models also produce overthinking errors, and the penalty reduces CoT length for full-precision models by 5.6\% to 10.0\% depending on the model. \textit{Quantization amplifies a failure mode already present in reasoning models}. Our KL divergence analysis is also not specific to quantization: any perturbation to a model's weights or activations, from pruning, distillation, or low-rank approximation, may similarly affect uncertain positions. Comparing token-level divergences between a reference model and its compressed variant could serve as a general diagnostic for identifying which aspects of generation are most vulnerable to compression.

Several directions remain open. Our penalty uses a fixed $\lambda$ at every decoding step. While the ablation shows robustness across $\lambda$ values, adapting $\lambda$ dynamically based on local entropy is a natural extension. The overthinking markers list is manually curated for English reasoning models. Automatically constructing the list per model would make our approach more widely applicable. Finally, our evaluation focuses on math, coding, and science benchmarks. Whether the same overthinking patterns arise in other reasoning and planning domains remains an open question.

\bibliographystyle{assets/plainnat}
\bibliography{paper}

\clearpage
\newpage
\beginappendix

\section{Experimental Details}
\label{app:exp-details}

\subsection{Calibration Data}

GPTQ is calibrated on \emph{reasoning data} to mitigate domain shift: we use model-generated reasoning traces on NuminaMath-1.5 \citep{li2024numinamath}, sampling 128 sequences of length 2048.
For AWQ, we follow the standard AWQ recipe and sample 128 sequences of length 512 from the Pile uncopyrighted dataset\footnote{\url{https://huggingface.co/datasets/monology/pile-uncopyrighted}} \citep{gao2020pile, liu2025quantization}.
For FlatQuant, we use 128 text sequences of length 2048 from WikiText2 \citep{merity2016pointer} by default. Following \citet{liu2025quantization}, we increase the calibration sequence length to 4096 for DeepSeek-R1-Distill-Qwen-1.5B and 7B to avoid extreme outliers in the K cache.

\subsection{Quantization Methods}

\paragraph{Asymmetric group-wise quantization.}
We use group-wise asymmetric quantization with group size $g=128$ along the input dimension, which is standard in GPTQ and AWQ quantization.
For each output channel $i$ and each consecutive group of $g$ weights within that channel, we quantize a weight vector $w \in \mathbb{R}^{g}$ as $\widehat{w} = s\,(q - z)$, where
$
q = \mathrm{clip}\!\left(\left\lfloor \frac{w}{s} + z \right\rceil,\; 0,\; 2^b-1 \right),
$
$b\in\{4,3\}$ is the bit-width, $s>0$ is a learned per-group scale, $z\in\mathbb{Z}$ is a per-group zero-point, $\lfloor \cdot \rceil$ denotes rounding to the nearest integer, and $\mathrm{clip}$ clamps to the representable integer range.

\paragraph{GPTQ (weight-only, 3/4-bit).}
GPTQ constructs $\widehat{W}$ by quantizing one layer at a time while using second-order information to compensate for quantization error \citep{frantar2022gptq}. Given calibration activations $X$, the goal is to minimize the output reconstruction error:
$
\min_{\widetilde{W}\in\mathcal{Q}_b} \;\; \|X\widetilde{W}^\top - XW^\top\|_F^2,
$
where $\mathcal{Q}_b$ denotes the set of matrices representable by the chosen $b$-bit group-wise quantizer. GPTQ approximately solves this using a blockwise sequential procedure and an approximate inverse-Hessian estimated from calibration data, implemented efficiently via Cholesky-based updates \citep{frantar2022gptq}. We use group size $g=128$ and evaluate 3-bit and 4-bit weight-only GPTQ.

\paragraph{AWQ (weight-only, 3/4-bit).}
AWQ is \emph{activation-aware} weight-only PTQ: it uses calibration activations to identify salient channels and applies a channel-wise rescaling before quantization to reduce output distortion \citep{lin2024awq}. For a diagonal scaling matrix $S$ applied to the input features, the linear layer can be rewritten as
$
XW^\top = (XS)\,(W S^{-1})^\top.
$
AWQ chooses $S$ from calibration statistics so that the rescaled weights $WS^{-1}$ are easier to quantize while preserving the original function. The rescaling is folded into the stored weights, so AWQ remains weight-only at inference time. We use group size $g=128$ and evaluate 3-bit and 4-bit weight-only AWQ.

\paragraph{FlatQuant (W4A4KV4 and W8A8KV8).}
Weight-only PTQ reduces model footprint but does not address other inference bottlenecks that dominate long-form generation, such as activation memory, bandwidth and KV-cache storage. To study these effects, we additionally use FlatQuant \citep{sun2024flatquant}, which quantizes weights, activations, and the KV cache with lightweight calibration, enabling end-to-end low-precision inference without gradient-based retraining. We evaluate two configurations: W4A4KV4 (4-bit weights, activations, and KV cache) and W8A8KV8 (8-bit weights, activations, and KV cache).

\paragraph{Bit-widths.}
Our baseline models use BF16 for weights, activations, and the KV cache. For weight-only PTQ, we evaluate $b\in\{4,3\}$ for both GPTQ and AWQ, matching the low-bit settings studied in prior work on quantized reasoning models \citep{liu2025quantization}. For FlatQuant, we evaluate W4A4KV4 and W8A8KV8.

\subsection{Penalty Sweeps}
\label{app:penalty-sweep}

We use temperature $T{=}0.6$ and nucleus sampling top-$p{=}0.95$ as default decoding hyperparameters throughout all experiments, following previous work \citep{liu2025quantization}.

\paragraph{Full model sweep (V1 overthinking markers).}
We evaluate the overthinking markers penalty (50 tokens, listed in \Cref{app:thinking-tokens}) across all five models and all quantization configurations. We sweep the penalty strength over $\lambda \in \{0.5, 0.7, 1.0, 1.5, 2.0, 2.6, 3.0, 4.0\}$ and evaluate on all five benchmarks. In \Cref{tab:main_results}, we report the best accuracy after the sweep over $\lambda$, the corresponding CoT length, alongside the standard deviation of accuracy and CoT length across all $\lambda$ values.
\Cref{fig:ablation} shows the average change in performance over all datasets and quantization configurations for each value of $\lambda$.

\paragraph{Token list ablation (Qwen-1.5B).}
We compare four token lists of equal size (50 tokens each) on Qwen-1.5B across six quantization configurations (GPTQ W3, GPTQ W4, AWQ W3, AWQ W4, FlatQuant W4A4KV4, FlatQuant W8A8KV8) and five benchmarks. We sweep $\lambda \in \{0.5, 0.7, 1.0, 1.5, 2.0, 2.6, 3.0, 4.0\}$ for each list. The four lists are: overthinking markers (50 hand-curated hesitation and redirection tokens), high-KL tokens (top 50 highest average KL tokens between BF16 and AWQ W3 with frequency $\geq$ 50), random tokens (50 randomly selected tokens not in the other lists), and low-KL tokens (50 tokens with the lowest average KL divergence with frequency $\geq$ 50). All lists are provided in \Cref{app:thinking-tokens}. This ablation totals $4 \times 6 \times 5 \times 8 = 960$ penalty runs plus 30 baseline runs. Results are shown in \Cref{fig:ablation}.

\paragraph{Negative penalty sweep (Qwen-1.5B).}
To verify that the direction of the intervention matters, we run an additional sweep with negative $\lambda$ values symmetric to the positive sweep: $\lambda \in \{-0.5, -0.7, -1.0, -1.5, -2.0,$ $ -2.6, -3.0, -4.0\}$. Negative penalties \emph{boost} the logits of the target tokens rather than suppressing them. We run the same four token lists across the same six quantization configurations and five benchmarks on Qwen-1.5B, totaling an additional 960 runs. Results are reported in \Cref{app:add-results}.

\section{Additional Results}
\label{app:add-results}

\subsection{Qualitative Example of Overthinking from MATH-500 }

\Cref{fig:qual_overthinking} shows an actual MATH-500 prompt and the corresponding model outputs from DeepSeek-R1-Distill-Qwen-1.5B in BF16 versus the same model quantized to 3-bit with AWQ.
While the quantized model initially reaches the correct intermediate result, it then samples repeated thinking/path-forking tokens (e.g., ``wait'', ``but'') and branches into alternative reasoning paths, leading to excessive CoT length and a failure to terminate. This is a representative overthinking failure mode example in our error analysis.

\begin{figure*}[h!]
\centering
\small
\resizebox{\textwidth}{!}{%
\begin{tikzpicture}[font=\small\sffamily, remember picture]
\node[anchor=west] (qtitle) at (0,0) {\textcolor{red!80!black}{\Large ?}\ \  \textbf{Question}};
\node[anchor=west, align=left] (qtext) at (0,-0.8) {%
What's the largest eight-digit base 2 integer? Express your answer in base 10.
(Please reason step by step, \\ and put your final answer within \texttt{\textbackslash boxed\{\}}.)};
\node[anchor=west, align=left] (lh) at (0,-2.2) {%
\textbf{Ideal steps followed by the non-quantized model}\\[-2pt]
$\simeq$ \textbf{1124 tokens}};
\node[anchor=west, align=left] (rh) at (9.2,-2.2) {%
\textbf{vs.\ 3-bit quantized model:}\\[-2pt]
\textbf{$\rightarrow$ 5494 tokens}};

\node[
  anchor=north west,
  draw=black!55,
  line width=1.2pt,
  rounded corners=2pt,
  inner sep=10pt,
  minimum width=0.45\textwidth
] (L) at (0,-3.2) {%
\begin{minipage}{0.45\textwidth}
\vspace{2pt}
\small
\begin{enumerate}[leftmargin=1.4em, itemsep=4pt]
  \item An ``eight-digit base 2 integer'' means a binary number with exactly 8 digits.
  \item The largest such number is \colorbox{black!5}{$11111111_2$}.
  \item Convert to base 10: \\
  $
  11111111_2 = \sum_{k=0}^{7} 2^k = 128+64+32+16+8+4+2+1 = 255
  $
  \item Therefore, the largest eight-digit base 2 integer in base 10 is:
  \[
  \boxed{255}\
  \]
\end{enumerate}
\end{minipage}
};

\node[
  anchor=north west,
  draw=black!55,
  line width=1.2pt,
  rounded corners=2pt,
  inner sep=10pt,
  minimum width=0.48\textwidth
] (R) at (9.2,-3.2) {%
\begin{minipage}{0.46\textwidth}
\small
\textbf{Correct reasoning start:}\\[3pt]
\colorbox{black!5}{\parbox{0.98\linewidth}{%
``So, the largest eight-digit binary number is $11111111_2$. Each digit represents a power of two $\ldots$
$128+64+32+16+8+4+2+1=255$.''\ \ \cmark
}}\\[4pt]

\textbf{Consecutive path-forking and overthinking:}\\[3pt]
\colorbox{black!5}{\parbox{0.98\linewidth}{%
``Wait, but maybe `eight-digit' means something else? \ldots
Wait, maybe it means eight digits in base 10 restricted to 0s and 1s? \ldots
Wait, what if the phrasing is tricky?''\ \ \xmark
}}\\[8pt]

\xmark \ \ \textbf{Model never terminates!}

\end{minipage}
};


\end{tikzpicture}}
\caption{\textbf{Quantization-induced overthinking in a single trace.}
We show an actual question from MATH-500 and the responses of DeepSeek-R1-Distill-Qwen-2.5-
1.5B in BF16 versus the same model quantized to 3-bit with AWQ.
The BF16 model follows a short, concise solution, while a 3-bit quantized model reaches the correct reasoning early but then injects path-forking ``wait''-style tokens that trigger branching and can stall termination.}
\label{fig:qual_overthinking}
\end{figure*}

\subsection{Position of Thinking Tokens in the Next-token Probability Distribution}

We analyze where thinking/hesitation tokens appear in the model’s next-token probability distribution, and how this changes between high-entropy and low-entropy decoding positions. Concretely, for each generation step \(t\) we compute the next-token entropy \(H_t\), rank all vocabulary items by their predicted probability, and record whether at least one token from our thinking-token lexicon (defined in \Cref{app:thinking-tokens}) appears among the top-20 most probable next tokens. We then compare this top-20 inclusion rate at \emph{high-entropy positions} (\(H_t > \tau\)) versus \emph{low-entropy positions} (\(H_t \le \tau\)), where \(\tau\) is set per model to the 80th percentile of its entropy distribution.

\begin{figure}[h!]
    \centering
    \includegraphics[width=0.55\textwidth]{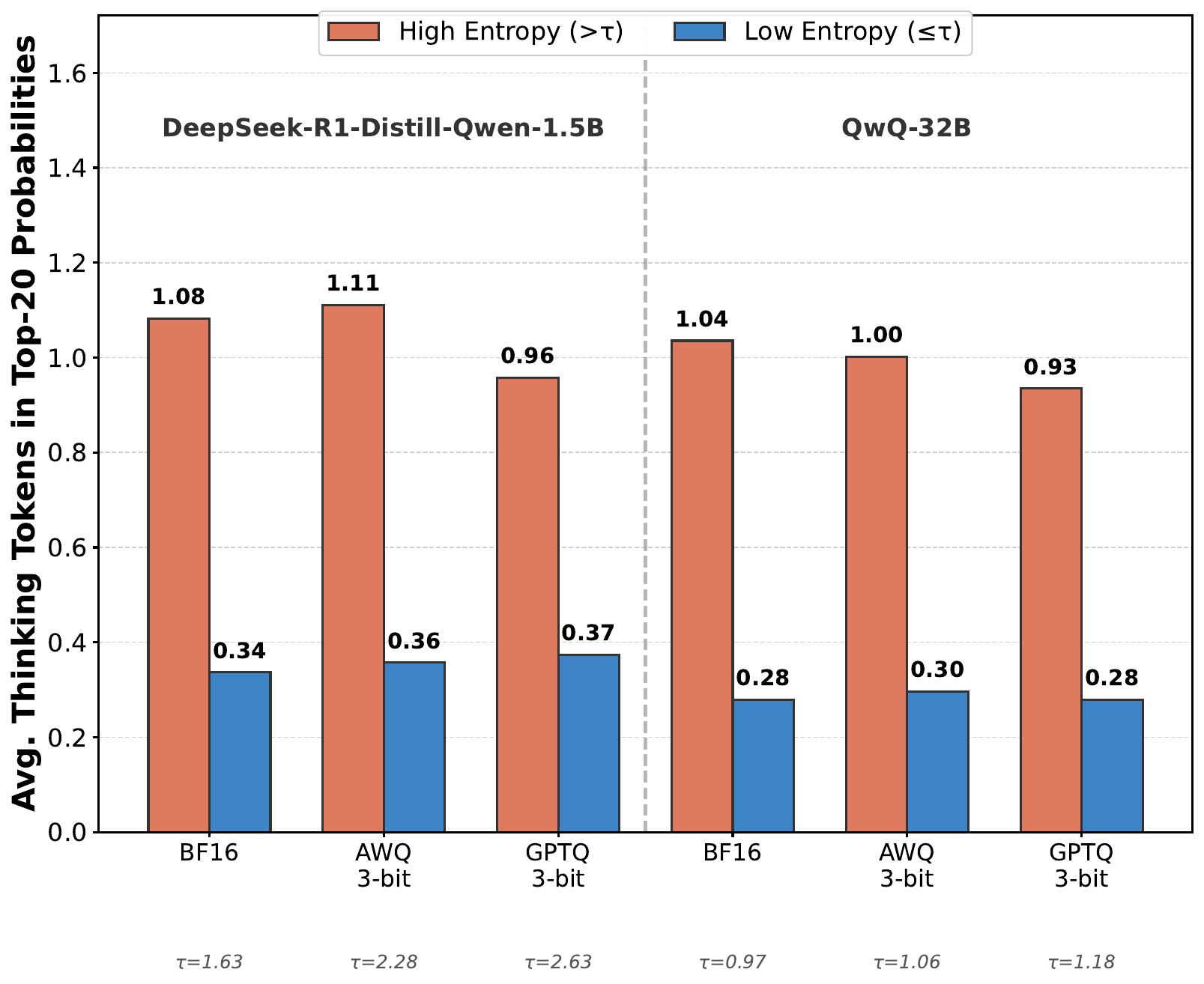}
\caption{\textbf{Thinking tokens are 2--4\(\times\) more likely to appear among the top-20 next-token predictions at high-entropy positions.}
At each decoding step, we measure whether a thinking token from the 50-token list in \Cref{app:tab:thinking-tokens} appears in the top-20 most probable next tokens, and compare high-entropy positions (entropy \(>\tau\)) to low-entropy positions (entropy \(\leq\tau\)).
We set \(\tau\) to the 80th percentile of the per-model entropy distribution similarly to \citet{wang2025beyond}, with $\tau$ being reported under each bar.
Across DeepSeek-R1-Distill-Qwen-1.5B and QwQ-32B, including their 3-bit quantized variants (AWQ and GPTQ), thinking tokens are substantially more likely to be ranked among the top predictions precisely when the model is locally uncertain.
Notably, these top-20 sampling rates of thinking tokens at high-entropy positions are high despite the large vocabulary sizes (151{,}936 for DeepSeek-R1-Distill-Qwen-1.5B; 152{,}064 for QwQ-32B), indicating that high-entropy positions concentrate probability mass onto tokens that can trigger overthinking.
Experiments are on MATH-500.}
    \label{app-fig:top-tokens}
\end{figure}

\Cref{app-fig:top-tokens} shows that thinking tokens are consistently \(\approx 2\)--\(4\times\) more likely to be ranked in the top-20 at high-entropy positions across DeepSeek-R1-Distill-Qwen-1.5B and QwQ-32B, including their 3-bit quantized variants (AWQ and GPTQ). Importantly, these inclusion rates are high despite the very large vocabulary sizes (151{,}936 tokens for DeepSeek-R1-Distill-Qwen-1.5B and 152{,}064 tokens for QwQ-32B). High-entropy positions are precisely the positions where thinking tokens become more likely under sampling, making them natural triggers for branching and overthinking.

\subsection{Word Cloud Visualizations of High-KL and Low-KL Tokens}
\label{app:wordclouds}

\begin{figure}[h]
\centering
\begin{subfigure}[t]{0.49\textwidth}
    \centering
    \includegraphics[width=\textwidth]{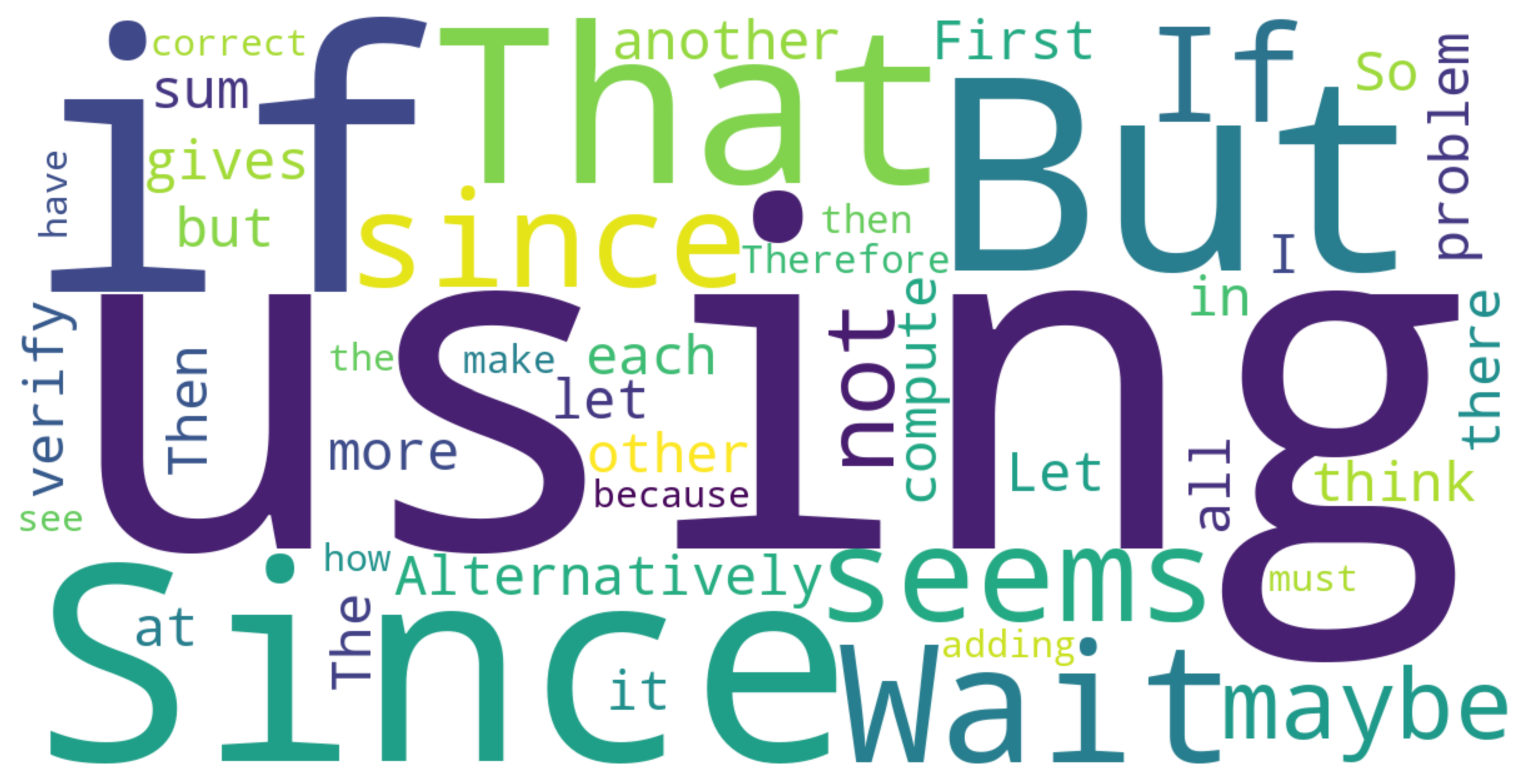}
    \caption{High-KL tokens (overthinking markers)}
    \label{fig:wc_high}
\end{subfigure}
\hfill
\begin{subfigure}[t]{0.49\textwidth}
    \centering
    \includegraphics[width=\textwidth]{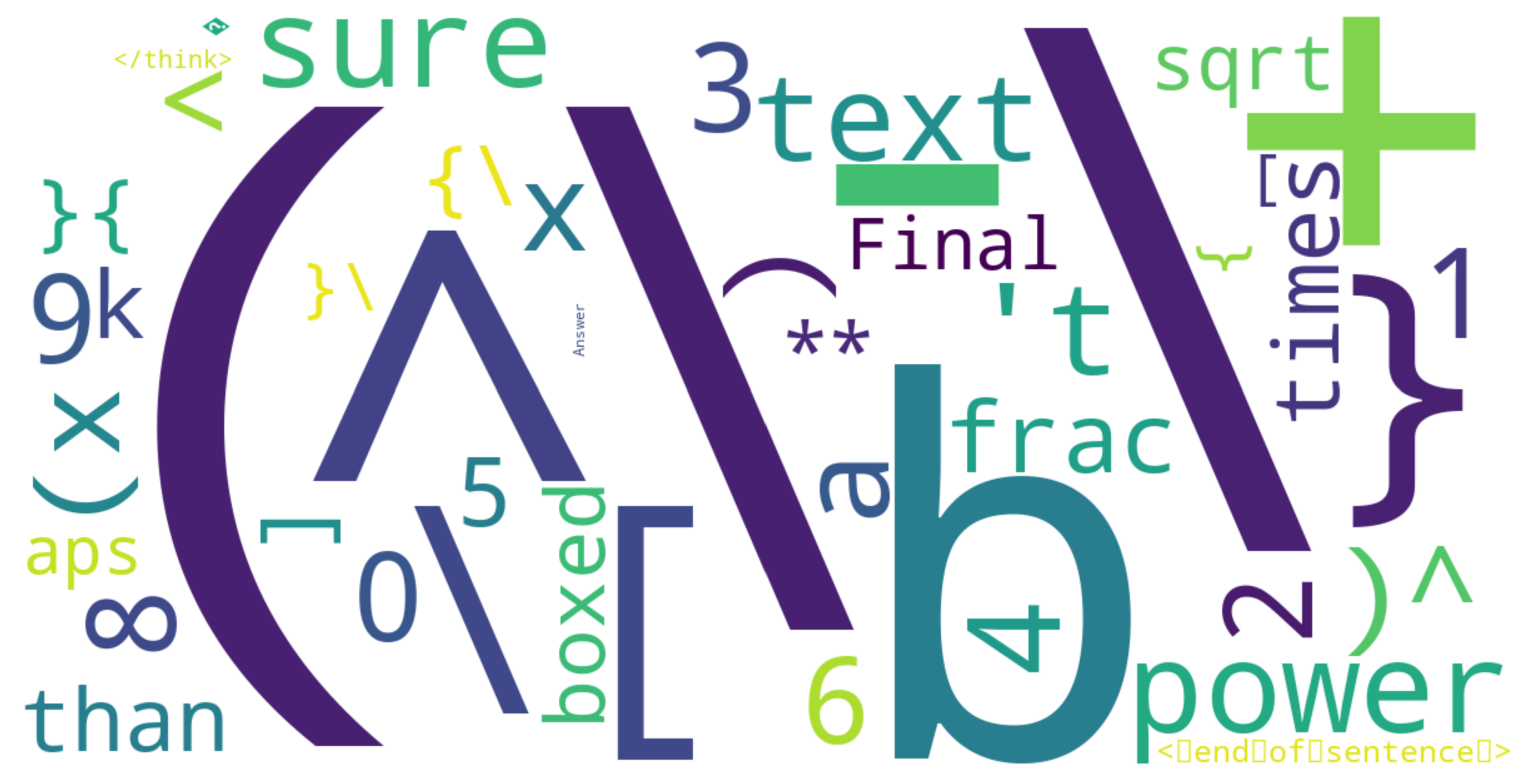}
    \caption{Low-KL tokens (math tokens)}
    \label{fig:wc_low}
\end{subfigure}
\caption{\textbf{Word clouds of the 50 highest-KL and 50 lowest-KL tokens between BF16 and AWQ 3-bit on MATH-500.} Token size reflects average KL divergence. High-KL tokens express thinking and branching. Low-KL tokens represent mathematical content.}
\label{fig:wordclouds}
\end{figure}

We show word cloud visualizations of the 50 tokens with the highest and lowest average KL divergence between the BF16 and 3-bit AWQ output distributions on MATH-500 in \Cref{fig:wordclouds}. Token size is proportional to the average KL divergence. The high-KL word cloud is dominated by hesitation and redirection tokens that open new reasoning branches, while the low-KL word cloud consists of mathematical symbols, digits, and formatting tokens that carry the computational content of the reasoning trace.

\subsection{Negative Penalty Sweep: Boosting Overthinking Markers}
\label{app:negative-sweep}

\begin{figure}[h!]
\centering
\begin{subfigure}[t]{0.49\textwidth}
    \centering
    \includegraphics[width=\textwidth]{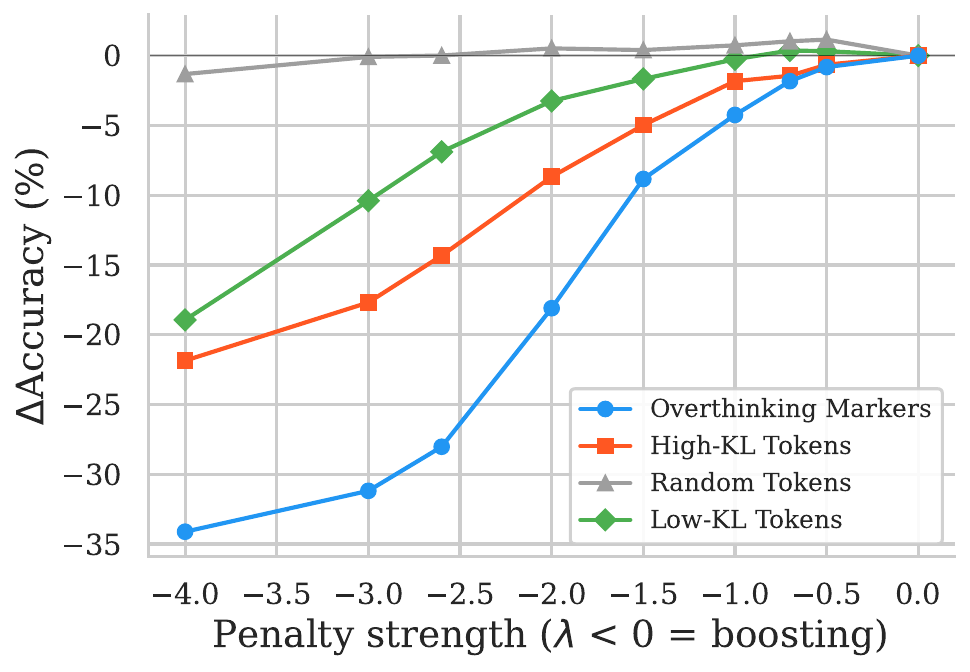} 
    \caption{Accuracy when boosting tokens}
    \label{fig:neg_acc}
\end{subfigure}
\hfill
\begin{subfigure}[t]{0.49\textwidth}
    \centering
    \includegraphics[width=\textwidth]{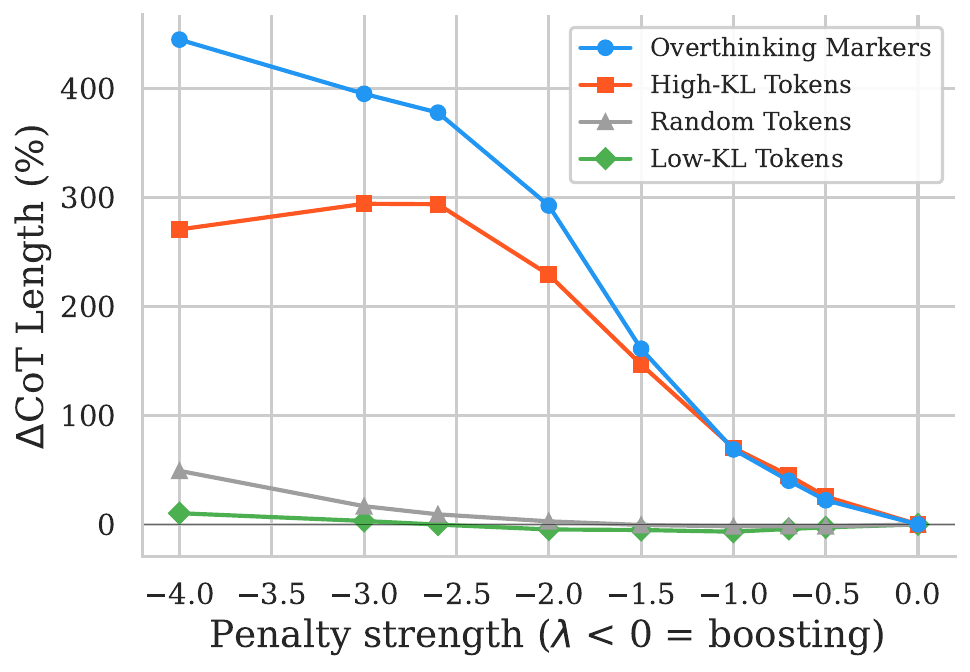}
    \caption{CoT length when boosting tokens}
    \label{fig:neg_len}
\end{subfigure}
\caption{\textbf{Boosting overthinking markers increases CoT length and degrades accuracy.} Negative $\lambda$ boosts rather than suppresses the target tokens. Overthinking markers and high-KL tokens cause severe CoT increases (up to +445\%) and accuracy drops (up to $-$34\%). Random and low-KL tokens have minimal effect on CoT length. Averaged across six quantization configurations and five benchmarks on Qwen-1.5B.}
\label{fig:neg_sweep}
\end{figure}

To verify that the direction of the penalty matters, we run an additional sweep with negative $\lambda$ values that boost rather than suppress the target tokens. \Cref{fig:neg_sweep} shows accuracy and CoT length changes when boosting each token list on Qwen-1.5B, averaged across all six quantization configurations and our five benchmarks. Boosting overthinking markers and high-KL tokens causes severe degradation in both accuracy and CoT length: at $\lambda = -3.0$, overthinking markers drop accuracy by 31\% while increasing CoT by 395\%, and high-KL tokens show a similar effect with an 18\% accuracy drop and 294\% CoT increase.
The similarity between these two lists further confirms that high-KL tokens are a good proxy for the tokens that drive overthinking under quantization. Random tokens remain nearly neutral across all negative $\lambda$ values, consistent with the positive $\lambda$ results. Low-KL tokens reduce accuracy at strong boost but barely increase CoT length, reaching only +10\% even at $\lambda = -4.0$. Combined with the positive $\lambda$ results in \Cref{fig:ablation}, these results confirm that overthinking markers and high-KL tokens are the ones that control CoT length: suppressing them reduces CoT, boosting them increases it.

\subsection{Error Analysis on GSM8K}
\label{app:error-gsm8k}

\begin{figure*}[h]
  \centering
  \begin{subfigure}[t]{0.49\textwidth}
    \centering
    \includegraphics[width=\linewidth]{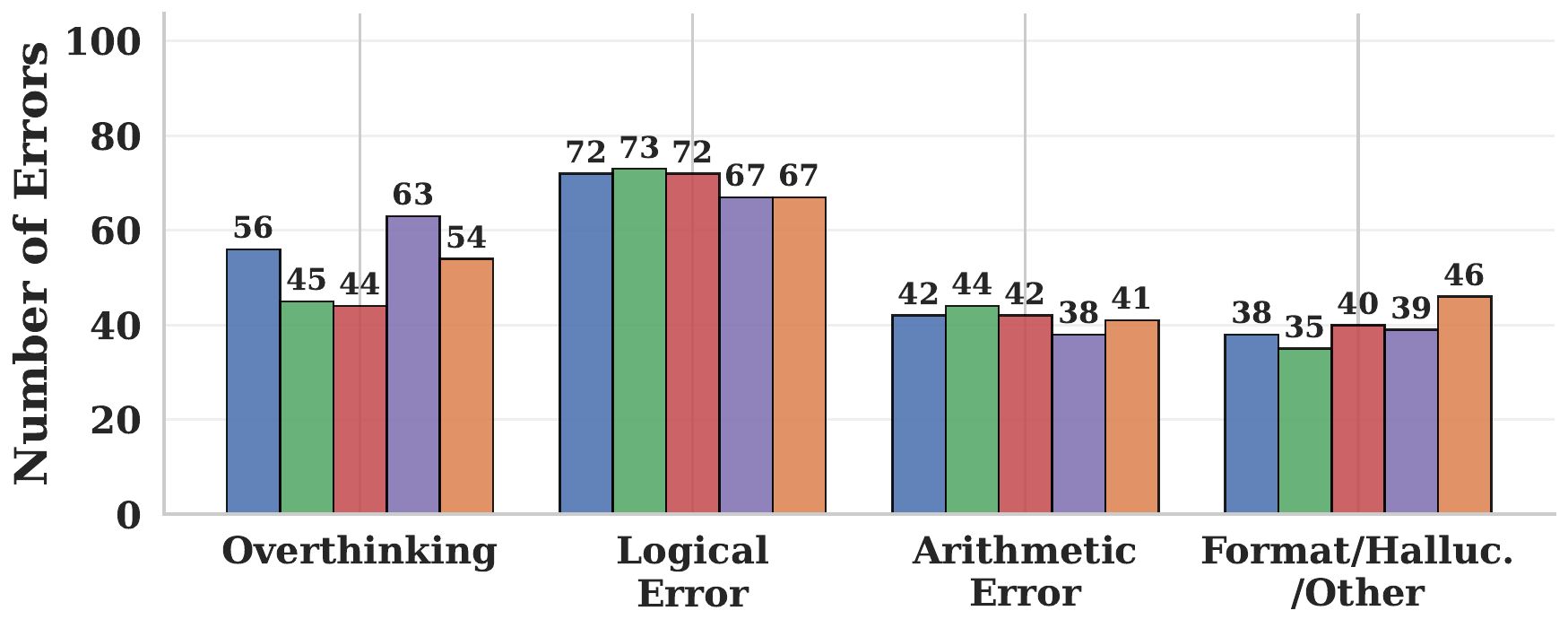}
    \caption{\small BF16}
  \end{subfigure}\hfill
  \begin{subfigure}[t]{0.49\textwidth}
    \centering
    \includegraphics[width=\linewidth]{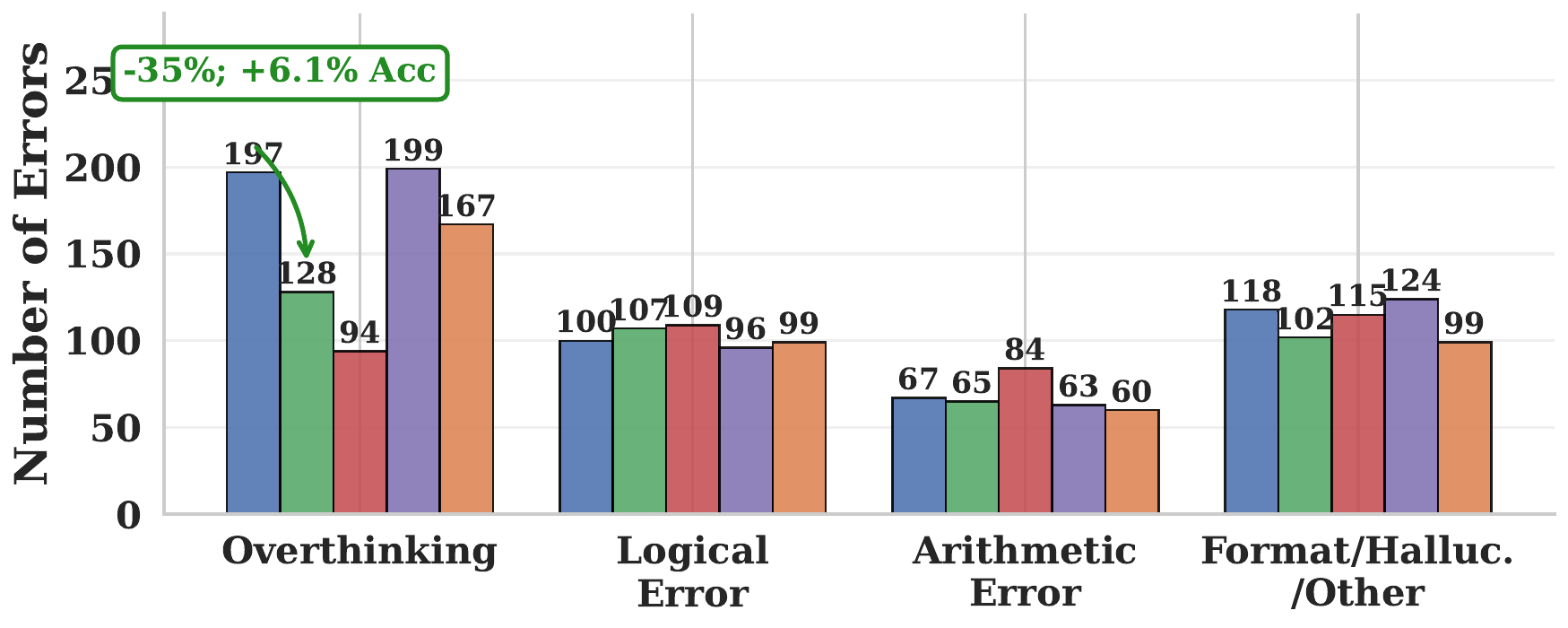}
    \caption{\small 3-bit AWQ}
  \end{subfigure}
  \begin{subfigure}[t]{0.49\textwidth}
    \centering
    \includegraphics[width=\linewidth]{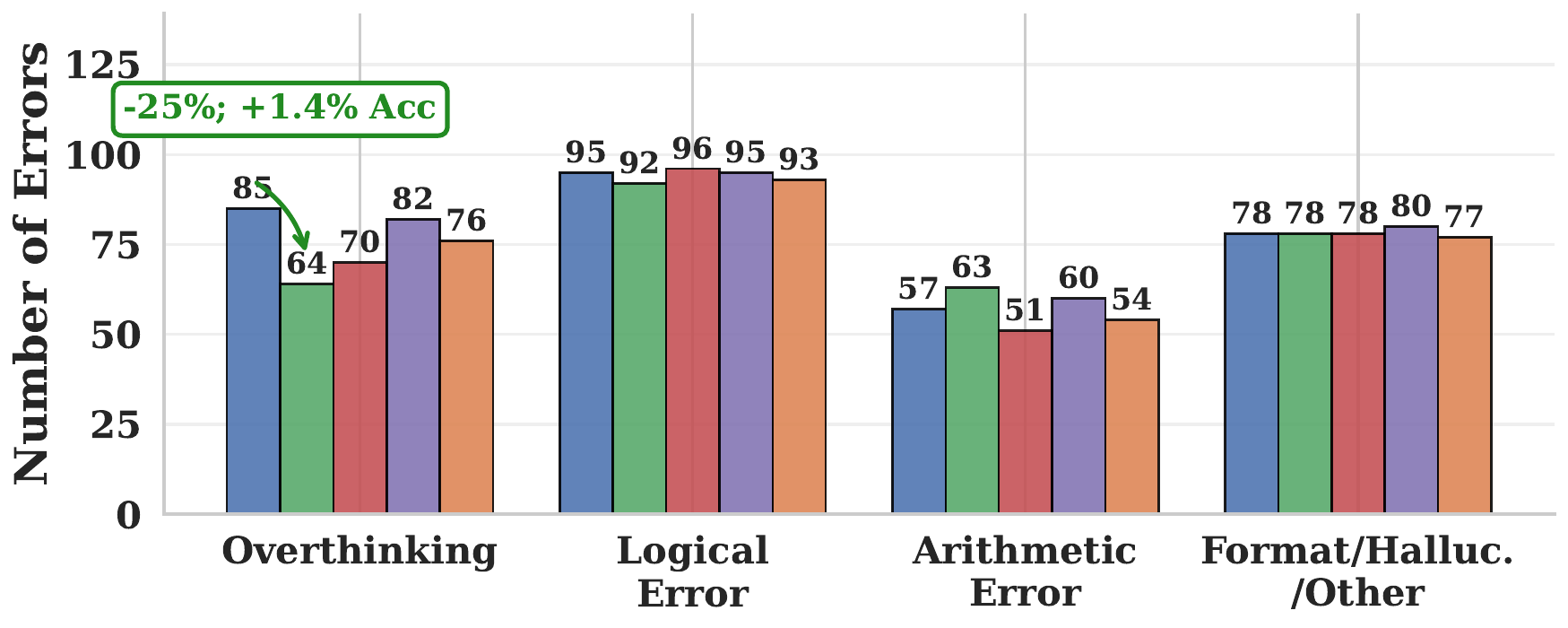}
    \caption{\small 3-bit GPTQ}
  \end{subfigure}\hfill
  \begin{subfigure}[t]{0.49\textwidth}
    \centering
    \includegraphics[width=\linewidth]{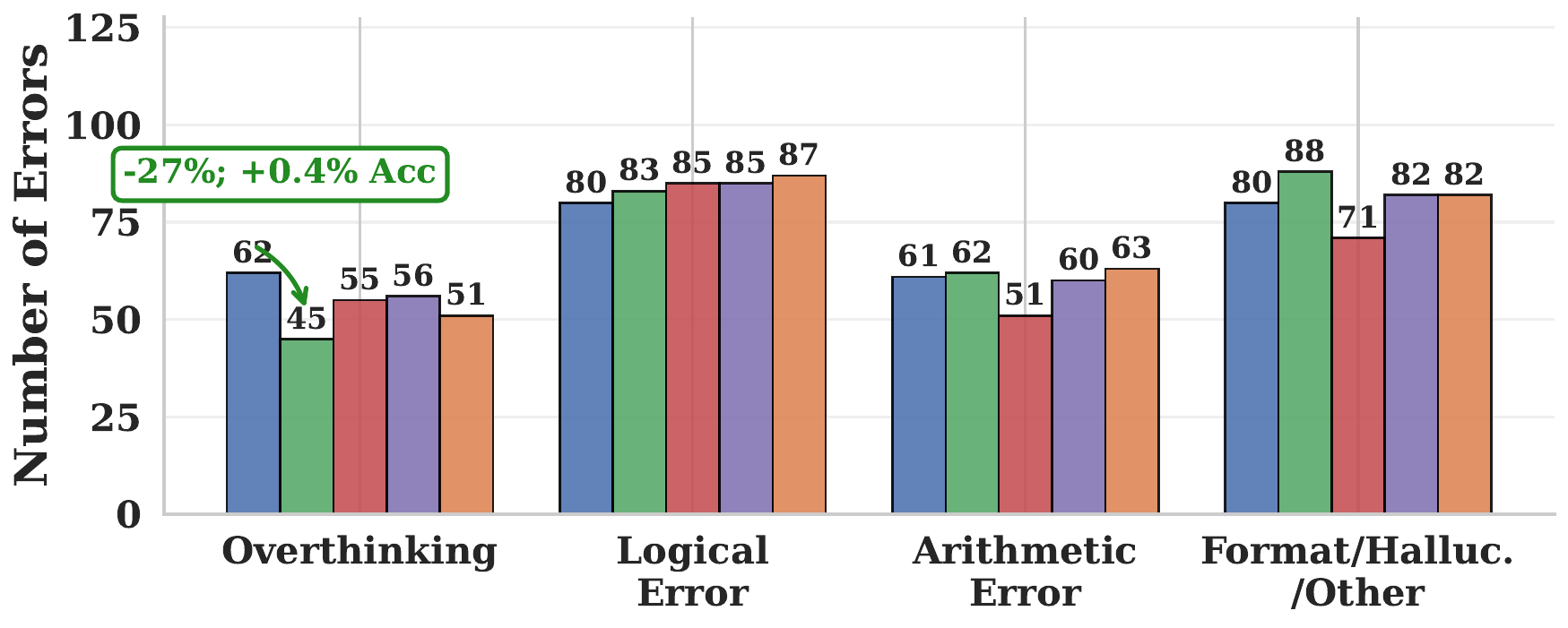}
    \caption{\small FlatQuant W4A4KV4}
  \end{subfigure}
  \includegraphics[width=0.89\textwidth]{figures/error_distribution_legend.pdf}
  \caption{\textbf{Error analysis on GSM8K confirms the overthinking pattern.} Same setup as \Cref{fig:error_breakdown} but on GSM8K. Penalizing overthinking markers reduces overthinking errors by up to 35\% on AWQ 3-bit while improving accuracy by 6.1\%.}
  \label{fig:error_gsm8k}
\end{figure*}

We repeat the error analysis from \Cref{sec:error-analysis} on GSM8K (1,319 questions) to verify that the overthinking pattern generalizes beyond MATH-500. \Cref{fig:error_gsm8k} shows the error breakdown for BF16 and three extreme quantization configurations. The same pattern holds: quantization increases overthinking errors, and penalizing overthinking markers reduces them. On AWQ 3-bit, overthinking errors drop from 197 to 128 ($-$35\%) while accuracy improves by 6.1\%.
As on MATH-500, penalizing random or low-KL tokens does not consistently reduce overthinking errors and can actually increase them.

\subsection{Full Quantization Evaluation}

We report a comprehensive evaluation of PTQ across model scales, datasets, and quantization methods in \Cref{app:tab:full_eval}. For each base model, we compare the BF16 checkpoint against multiple weight-only quantization settings (GPTQ, AWQ) at 4-bit and 3-bit precision, as well as FlatQuant configurations that quantize weights, activations, and the KV cache. We summarize results using two metrics: task accuracy (Acc) and average chain-of-thought length in tokens (Tok).

Across all models and benchmarks, we observe the same pattern: \textbf{more aggressive quantization generally decreases accuracy while increasing CoT length}. This effect is especially pronounced at 3-bit precision, where models often exhibit large CoT increases while simultaneously losing substantial accuracy. In contrast, 4-bit quantization is typically less disruptive: several W4A16KV16 configurations remain close to BF16 accuracy with modest CoT increases. FlatQuant at W8A8KV8 is the most stable setting overall, often matching BF16 accuracy with small CoT changes, whereas W4A4KV4 can reintroduce the same pattern of longer reasoning with degraded accuracy.

\begin{table}[h!]
\centering
\caption{\textbf{Effect of quantization on accuracy and CoT length across models and benchmarks.} We report accuracy (\%) and average CoT length (tokens) for BF16 and each quantization configuration. More aggressive quantization generally decreases accuracy while increasing CoT length.}
\label{app:tab:full_eval}
\resizebox{\textwidth}{!}{%
\scriptsize
\begin{tabular}{ll|cc|cc|cc|cc|cc|cc|cc}
\toprule
\multirow{2}{*}{\textbf{Model}} & \multirow{2}{*}{\textbf{Quantization}} & \multicolumn{2}{c|}{\textbf{MATH-500}} & \multicolumn{2}{c|}{\textbf{GSM8K}} & \multicolumn{2}{c|}{\textbf{AIME-120}} & \multicolumn{2}{c|}{\textbf{LiveCodeBench}} & \multicolumn{2}{c|}{\textbf{GPQA-Diamond}} & \multicolumn{2}{c|}{\textbf{Avg.}} & \multicolumn{2}{c}{\textbf{$\Delta$ vs BF16}} \\
& & Acc & Tok & Acc & Tok & Acc & Tok & Acc & Tok & Acc & Tok & Acc & Tok & Acc & Tok \\
\midrule
\multirow{7}{*}{Qwen 1.5B} & BF16 & 85.6 & 5156 & 84.2 & 2664 & 24.4 & 17465 & 17.5 & 19066 & 30.8 & 11887 & 48.5 & 11248 & 0.0 & 0 \\
 & GPTQ W4A16KV16 & 82.8 & 6192 & 82.4 & 3979 & 17.8 & 17770 & 12.7 & 20529 & 30.3 & 13202 & 45.2 & 12334 & \cellcolor{red!14}-3.3 & \cellcolor{red!12}+1087 \\
 & GPTQ W3A16KV16 & 71.6 & 6813 & 76.1 & 3996 & 8.9 & 20458 & 6.3 & 25854 & 26.3 & 18010 & 37.8 & 15026 & \cellcolor{red!25}-10.7 & \cellcolor{red!17}+3779 \\
 & AWQ W4A16KV16 & 83.0 & 6321 & 83.5 & 3732 & 17.8 & 18123 & 12.7 & 23711 & 34.3 & 13422 & 46.3 & 13062 & \cellcolor{red!13}-2.2 & \cellcolor{red!13}+1814 \\
 & AWQ W3A16KV16 & 47.0 & 23444 & 63.5 & 16263 & 4.4 & 33015 & 4.5 & 38793 & 25.8 & 28726 & 29.0 & 28048 & \cellcolor{red!35}-19.5 & \cellcolor{red!35}+16801 \\
 & FlatQuant W8A8KV8 & 84.8 & 5008 & 84.3 & 2719 & 22.2 & 17727 & 17.2 & 19947 & 36.9 & 12083 & 49.1 & 11497 & \cellcolor{green!11}+0.6 & \cellcolor{red!10}+249 \\
 & FlatQuant W4A4KV4 & 66.4 & 8041 & 78.5 & 3409 & 7.8 & 17515 & 7.5 & 25383 & 30.8 & 20813 & 38.2 & 15032 & \cellcolor{red!25}-10.3 & \cellcolor{red!17}+3785 \\
\midrule
\multirow{7}{*}{Qwen 7B} & BF16 & 93.2 & 3937 & 91.1 & 1882 & 45.6 & 12733 & 37.7 & 16478 & 50.0 & 9911 & 63.5 & 8988 & 0.0 & 0 \\
 & GPTQ W4A16KV16 & 92.6 & 4584 & 91.1 & 2233 & 45.6 & 14306 & 38.4 & 16048 & 58.1 & 10903 & 65.2 & 9615 & \cellcolor{green!14}+1.6 & \cellcolor{red!11}+627 \\
 & GPTQ W3A16KV16 & 91.0 & 4831 & 90.1 & 2525 & 30.0 & 16969 & 27.2 & 19139 & 41.9 & 11417 & 56.0 & 10976 & \cellcolor{red!21}-7.5 & \cellcolor{red!13}+1988 \\
 & AWQ W4A16KV16 & 93.4 & 3977 & 90.6 & 1959 & 43.3 & 14359 & 36.2 & 15787 & 47.5 & 10636 & 62.2 & 9344 & \cellcolor{red!11}-1.3 & \cellcolor{red!10}+355 \\
 & AWQ W3A16KV16 & 91.2 & 4842 & 90.4 & 2142 & 30.0 & 16148 & 26.9 & 21288 & 48.0 & 10656 & 57.3 & 11015 & \cellcolor{red!19}-6.2 & \cellcolor{red!14}+2027 \\
 & FlatQuant W8A8KV8 & 93.0 & 3840 & 90.4 & 1919 & 40.0 & 13412 & 35.4 & 15526 & 49.5 & 9880 & 61.7 & 8915 & \cellcolor{red!12}-1.9 & -73 \\
 & FlatQuant W4A4KV4 & 83.6 & 5444 & 87.7 & 2120 & 23.3 & 20409 & 12.3 & 23862 & 50.0 & 17957 & 51.4 & 13958 & \cellcolor{red!28}-12.1 & \cellcolor{red!19}+4970 \\
\midrule
\multirow{7}{*}{Qwen 14B} & BF16 & 94.6 & 3735 & 94.1 & 1734 & 57.8 & 13116 & 49.6 & 14435 & 57.6 & 9489 & 70.7 & 8502 & 0.0 & 0 \\
 & GPTQ W4A16KV16 & 95.6 & 3437 & 93.3 & 1641 & 62.2 & 11193 & 52.6 & 15313 & 60.1 & 9387 & 72.8 & 8194 & \cellcolor{green!16}+2.0 & \cellcolor{green!10}-308 \\
 & GPTQ W3A16KV16 & 94.4 & 3607 & 92.8 & 1592 & 44.4 & 12397 & 42.2 & 14643 & 53.5 & 9130 & 65.5 & 8274 & \cellcolor{red!17}-5.3 & \cellcolor{green!10}-228 \\
 & AWQ W4A16KV16 & 95.0 & 3626 & 92.6 & 1646 & 56.7 & 12426 & 48.1 & 14751 & 59.6 & 9340 & 70.4 & 8358 & -0.3 & \cellcolor{green!10}-144 \\
 & AWQ W3A16KV16 & 93.0 & 4277 & 93.0 & 2064 & 48.9 & 14331 & 42.2 & 17533 & 56.6 & 10458 & 66.7 & 9733 & \cellcolor{red!16}-4.0 & \cellcolor{red!12}+1231 \\
 & FlatQuant W8A8KV8 & 94.8 & 3826 & 94.2 & 1728 & 61.1 & 11895 & 51.9 & 14008 & 57.6 & 9402 & 71.9 & 8172 & \cellcolor{green!13}+1.2 & \cellcolor{green!10}-330 \\
 & FlatQuant W4A4KV4 & 95.0 & 3675 & 93.5 & 1802 & 52.2 & 12667 & 49.3 & 15221 & 63.6 & 8752 & 70.7 & 8423 & -0.0 & -78 \\
\midrule
\multirow{7}{*}{Llama 8B} & BF16 & 89.0 & 4629 & 89.0 & 2192 & 44.4 & 14533 & 35.1 & 16176 & 49.0 & 10225 & 61.3 & 9551 & 0.0 & 0 \\
 & GPTQ W4A16KV16 & 89.6 & 4640 & 88.6 & 2327 & 41.1 & 14123 & 35.1 & 15467 & 48.0 & 10727 & 60.5 & 9457 & \cellcolor{red!11}-0.8 & -94 \\
 & GPTQ W3A16KV16 & 82.2 & 6181 & 85.4 & 2362 & 26.7 & 18230 & 24.6 & 24081 & 29.8 & 12819 & 49.7 & 12735 & \cellcolor{red!27}-11.6 & \cellcolor{red!16}+3184 \\
 & AWQ W4A16KV16 & 89.2 & 4763 & 88.0 & 2206 & 30.0 & 15470 & 35.1 & 16893 & 48.0 & 10458 & 58.1 & 9958 & \cellcolor{red!14}-3.2 & \cellcolor{red!10}+407 \\
 & AWQ W3A16KV16 & 80.4 & 5357 & 82.6 & 2428 & 12.2 & 14787 & 23.1 & 18136 & 35.9 & 9631 & 46.8 & 10068 & \cellcolor{red!31}-14.5 & \cellcolor{red!11}+517 \\
 & FlatQuant W8A8KV8 & 88.8 & 4595 & 89.6 & 2144 & 37.8 & 14410 & 37.3 & 15641 & 48.5 & 10611 & 60.4 & 9480 & \cellcolor{red!11}-0.9 & -71 \\
 & FlatQuant W4A4KV4 & 86.4 & 4988 & 87.6 & 2355 & 22.2 & 15405 & 35.1 & 17833 & 40.9 & 10239 & 54.4 & 10164 & \cellcolor{red!20}-6.9 & \cellcolor{red!11}+613 \\
\midrule
\multirow{5}{*}{QwQ 32B} & BF16 & 97.2 & 4494 & 95.7 & 1913 & 76.7 & 15345 & 62.3 & 17914 & 63.1 & 10071 & 79.0 & 9947 & 0.0 & 0 \\
 & GPTQ W4A16KV16 & 97.4 & 4450 & 95.6 & 1855 & 78.9 & 15710 & 58.6 & 19034 & 59.6 & 9914 & 78.0 & 10193 & \cellcolor{red!11}-1.0 & \cellcolor{red!10}+245 \\
 & GPTQ W3A16KV16 & 96.8 & 4902 & 95.4 & 1973 & 64.4 & 16073 & 50.4 & 19090 & 56.1 & 9252 & 72.6 & 10258 & \cellcolor{red!19}-6.4 & \cellcolor{red!10}+311 \\
 & AWQ W4A16KV16 & 97.4 & 4486 & 95.8 & 1936 & 75.6 & 15733 & 59.0 & 19111 & 62.6 & 10767 & 78.1 & 10407 & \cellcolor{red!11}-0.9 & \cellcolor{red!10}+459 \\
 & AWQ W3A16KV16 & 96.0 & 4612 & 95.9 & 1815 & 60.0 & 16687 & 51.5 & 19493 & 61.1 & 10121 & 72.9 & 10546 & \cellcolor{red!19}-6.1 & \cellcolor{red!11}+598 \\
\bottomrule
\end{tabular}%
}
\end{table}

\subsection{Effect of the Overthinking Penalty}
\label{app:penalty-tables}

We report the effect of the overthinking penalty for each model individually in \Cref{app:tab:qwen15b,app:tab:qwen7b,app:tab:qwen14b,app:tab:llama8b,app:tab:qwq32b}. For each quantization configuration, we show baseline accuracy and CoT length (Base row) and the results with the best penalty $\lambda$ (+Pen row). The $\Delta$ columns report the mean change across the five benchmarks with standard deviation across all $\lambda$ values in the sweep.

The penalty consistently reduces CoT length across all models and quantization configurations. The CoT reduction is largest for the most aggressively quantized configurations (3-bit weight-only and W4A4KV4), where overthinking is most severe, and smallest for mild quantization (4-bit weight-only and W8A8KV8). Accuracy is preserved or improved in the majority of configurations. The few cases where accuracy slightly decreases still show substantial CoT reductions, yielding a net improvement in the efficiency-accuracy tradeoff. Larger models (Qwen-14B, QwQ-32B) show smaller absolute CoT reductions but the penalty remains effective, confirming that the intervention generalizes across model scales.

\begin{table}[h]
\centering
\caption{\textbf{Effect of the overthinking penalty on Qwen-1.5B.} For each quantization configuration, the top row (Base) shows accuracy (\%) and CoT length (thousands of tokens) without penalty, and the bottom row (+Pen) shows results with the best penalty $\lambda$. Per-benchmark cells are colored when the penalty changes accuracy or CoT length by $\geq$2\%. $\Delta$Acc and $\Delta$Len\% report the mean change across the five benchmarks with std across $\lambda$ values.}
\label{app:tab:qwen15b}
\resizebox{\textwidth}{!}{%
\setlength{\tabcolsep}{3.5pt}
\renewcommand{\arraystretch}{1.0}
\scriptsize
\begin{tabular}{cl@{\hskip 4pt}l ccccc c cc}
\toprule
 & &  & \textbf{AIME} & \textbf{MATH} & \textbf{GSM8K} & \textbf{GPQA} & \textbf{LCB} & \textbf{Avg.} & \textbf{$\Delta$Acc} & \textbf{$\Delta$Len\%} \\
\cmidrule(lr){10-10} \cmidrule(lr){11-11}
 & &  & \multicolumn{5}{c}{\footnotesize\textit{Acc (\%)\;/\;Len (k)}} & \footnotesize\textit{Acc\;/\;Len} & \multicolumn{2}{c}{\footnotesize\textit{mean$_{\pm\text{std}}$}} \\
\midrule
\multirow{14}{*}{\rotatebox{90}{Qwen-1.5B}} & \multirow{2}{*}{\textbf{BF16}} & Base & \textbf{24.4}\,/\,\textbf{17.5} & \textbf{85.6}\,/\,\textbf{5.2} & \textbf{84.2}\,/\,\textbf{2.7} & \textbf{30.8}\,/\,\textbf{11.9} & \textbf{17.5}\,/\,\textbf{19.1} & \textbf{48.5}\,/\,\textbf{11.2} & & \\
 & & +Pen & 25.6\,/\,\colorbox{green!17}{14.8} & 85.8\,/\,\colorbox{green!15}{4.5} & 85.1\,/\,\colorbox{green!9}{2.6} & \colorbox{green!35}{39.4}\,/\,\colorbox{green!14}{10.5} & 17.5\,/\,\colorbox{green!10}{18.3} & 50.7\,/\,10.1 & \cellcolor{green!24}$+2.1_{\scriptscriptstyle\pm 1.6}$ & \cellcolor{green!19}$-9.2_{\scriptscriptstyle\pm 5.7}$ \\
\cline{2-11}
 & \multirow{2}{*}{GPTQ W4} & Base & 17.8\,/\,17.8 & 82.8\,/\,6.2 & 82.4\,/\,4.0 & 30.3\,/\,13.2 & 12.7\,/\,20.5 & 45.2\,/\,12.3 & & \\
 & & +Pen & \colorbox{green!29}{24.4}\,/\,\colorbox{green!16}{15.3} & \colorbox{green!16}{84.8}\,/\,\colorbox{green!21}{4.8} & 83.2\,/\,\colorbox{green!25}{2.8} & \colorbox{green!23}{34.8}\,/\,\colorbox{green!19}{10.7} & 14.6\,/\,\colorbox{green!20}{16.4} & 48.4\,/\,10.0 & \cellcolor{green!32}$+3.2_{\scriptscriptstyle\pm 1.2}$ & \cellcolor{green!33}$-20.9_{\scriptscriptstyle\pm 7.2}$ \\
\cline{2-11}
 & \multirow{2}{*}{GPTQ W3} & Base & 8.9\,/\,20.5 & 71.6\,/\,6.8 & 76.1\,/\,4.0 & 26.3\,/\,18.0 & 6.3\,/\,25.9 & 37.8\,/\,15.0 & & \\
 & & +Pen & \colorbox{green!16}{11.1}\,/\,\colorbox{green!22}{15.6} & 73.0\,/\,\colorbox{green!16}{5.8} & 77.5\,/\,\colorbox{green!26}{2.8} & \colorbox{green!17}{28.8}\,/\,\colorbox{green!26}{12.5} & 8.2\,/\,\colorbox{green!16}{22.2} & 39.7\,/\,11.8 & \cellcolor{green!23}$+1.9_{\scriptscriptstyle\pm 1.1}$ & \cellcolor{green!35}$-22.6_{\scriptscriptstyle\pm 6.2}$ \\
\cline{2-11}
 & \multirow{2}{*}{AWQ W4} & Base & 17.8\,/\,18.1 & 83.0\,/\,6.3 & 83.5\,/\,3.7 & 34.3\,/\,13.4 & 12.7\,/\,23.7 & 46.3\,/\,13.1 & & \\
 & & +Pen & \colorbox{green!26}{23.3}\,/\,\colorbox{green!12}{16.7} & 84.0\,/\,\colorbox{green!21}{4.9} & 84.2\,/\,\colorbox{green!16}{3.2} & \colorbox{green!22}{38.4}\,/\,\colorbox{green!14}{12.0} & \colorbox{green!16}{14.9}\,/\,\colorbox{green!18}{19.4} & 49.0\,/\,11.2 & \cellcolor{green!29}$+2.7_{\scriptscriptstyle\pm 1.5}$ & \cellcolor{green!25}$-14.8_{\scriptscriptstyle\pm 5.2}$ \\
\cline{2-11}
 & \multirow{2}{*}{AWQ W3} & Base & 4.4\,/\,33.0 & 47.0\,/\,23.4 & 63.5\,/\,16.3 & 25.8\,/\,28.7 & 4.5\,/\,38.8 & 29.0\,/\,28.0 & & \\
 & & +Pen & \colorbox{green!23}{8.9}\,/\,\colorbox{green!14}{29.3} & \colorbox{green!40}{61.2}\,/\,\colorbox{green!35}{12.9} & \colorbox{green!28}{69.5}\,/\,\colorbox{green!35}{7.6} & \colorbox{green!23}{30.3}\,/\,\colorbox{green!20}{22.7} & 6.3\,/\,\colorbox{green!13}{35.3} & 35.2\,/\,21.6 & \cellcolor{green!45}$+6.2_{\scriptscriptstyle\pm 1.7}$ & \cellcolor{green!41}$-28.0_{\scriptscriptstyle\pm 7.5}$ \\
\cline{2-11}
 & \multirow{2}{*}{FQ W8A8} & Base & 22.2\,/\,17.7 & 84.8\,/\,5.0 & 84.3\,/\,2.7 & 36.9\,/\,12.1 & 17.2\,/\,19.9 & 49.1\,/\,11.5 & & \\
 & & +Pen & \colorbox{green!23}{26.7}\,/\,\colorbox{green!16}{15.3} & 85.4\,/\,\colorbox{green!12}{4.6} & 85.1\,/\,\colorbox{green!18}{2.2} & \colorbox{green!16}{38.9}\,/\,\colorbox{green!15}{10.6} & 15.7\,/\,\colorbox{green!12}{18.4} & 50.4\,/\,10.2 & \cellcolor{green!18}$+1.3_{\scriptscriptstyle\pm 1.5}$ & \cellcolor{green!22}$-11.8_{\scriptscriptstyle\pm 4.5}$ \\
\cline{2-11}
 & \multirow{2}{*}{FQ W4A4} & Base & 7.8\,/\,17.5 & 66.4\,/\,8.0 & 78.5\,/\,3.4 & 30.8\,/\,20.8 & 7.5\,/\,25.4 & 38.2\,/\,15.0 & & \\
 & & +Pen & \colorbox{green!23}{12.2}\,/\,\colorbox{green!10}{16.8} & \colorbox{green!25}{71.4}\,/\,\colorbox{green!21}{6.2} & 78.9\,/\,\colorbox{green!31}{2.1} & 31.3\,/\,\colorbox{green!17}{17.6} & 8.6\,/\,\colorbox{green!11}{23.8} & 40.5\,/\,13.3 & \cellcolor{green!25}$+2.3_{\scriptscriptstyle\pm 1.3}$ & \cellcolor{green!29}$-17.6_{\scriptscriptstyle\pm 4.9}$ \\
\bottomrule
\end{tabular}%
}
\end{table}

\begin{table}[h]
\centering
\caption{\textbf{Effect of the overthinking penalty on Qwen-7B.} For each quantization configuration, the top row (Base) shows accuracy (\%) and CoT length (thousands of tokens) without penalty, and the bottom row (+Pen) shows results with the best penalty $\lambda$. Per-benchmark cells are colored when the penalty changes accuracy or CoT length by $\geq$2\%. $\Delta$Acc and $\Delta$Len\% report the mean change across the five benchmarks with std across $\lambda$ values.}
\label{app:tab:qwen7b}
\resizebox{\textwidth}{!}{%
\setlength{\tabcolsep}{3.5pt}
\renewcommand{\arraystretch}{1.0}
\scriptsize
\begin{tabular}{cl@{\hskip 4pt}l ccccc c cc}
\toprule
 & &  & \textbf{AIME} & \textbf{MATH} & \textbf{GSM8K} & \textbf{GPQA} & \textbf{LCB} & \textbf{Avg.} & \textbf{$\Delta$Acc} & \textbf{$\Delta$Len\%} \\
\cmidrule(lr){10-10} \cmidrule(lr){11-11}
 & &  & \multicolumn{5}{c}{\footnotesize\textit{Acc (\%)\;/\;Len (k)}} & \footnotesize\textit{Acc\;/\;Len} & \multicolumn{2}{c}{\footnotesize\textit{mean$_{\pm\text{std}}$}} \\
\midrule
\multirow{14}{*}{\rotatebox{90}{Qwen-7B}} & \multirow{2}{*}{\textbf{BF16}} & Base & \textbf{45.6}\,/\,\textbf{12.7} & \textbf{93.2}\,/\,\textbf{3.9} & \textbf{91.1}\,/\,\textbf{1.9} & \textbf{50.0}\,/\,\textbf{9.9} & \textbf{37.7}\,/\,\textbf{16.5} & \textbf{63.5}\,/\,\textbf{9.0} & & \\
 & & +Pen & \colorbox{green!29}{52.2}\,/\,\colorbox{green!15}{11.2} & 94.2\,/\,\colorbox{green!14}{3.5} & 91.6\,/\,\colorbox{green!15}{1.6} & \colorbox{green!29}{56.6}\,/\,\colorbox{green!10}{9.4} & 38.8\,/\,\colorbox{green!13}{14.9} & 66.7\,/\,8.1 & \cellcolor{green!32}$+3.2_{\scriptscriptstyle\pm 1.9}$ & \cellcolor{green!20}$-10.0_{\scriptscriptstyle\pm 2.7}$ \\
\cline{2-11}
 & \multirow{2}{*}{GPTQ W4} & Base & 45.6\,/\,14.3 & 92.6\,/\,4.6 & 91.1\,/\,2.2 & 58.1\,/\,10.9 & 38.4\,/\,16.0 & 65.2\,/\,9.6 & & \\
 & & +Pen & 46.7\,/\,\colorbox{green!12}{13.3} & 93.4\,/\,\colorbox{green!17}{3.8} & 91.5\,/\,\colorbox{green!17}{1.9} & \colorbox{red!29}{51.5}\,/\,\colorbox{green!9}{10.6} & \colorbox{red!17}{35.8}\,/\,\colorbox{green!10}{15.4} & 63.8\,/\,9.0 & \cellcolor{red!19}$-1.4_{\scriptscriptstyle\pm 1.7}$ & \cellcolor{green!19}$-9.3_{\scriptscriptstyle\pm 2.8}$ \\
\cline{2-11}
 & \multirow{2}{*}{GPTQ W3} & Base & 30.0\,/\,17.0 & 91.0\,/\,4.8 & 90.1\,/\,2.5 & 41.9\,/\,11.4 & 27.2\,/\,19.1 & 56.0\,/\,11.0 & & \\
 & & +Pen & \colorbox{green!23}{34.4}\,/\,\colorbox{green!15}{14.8} & 91.2\,/\,\colorbox{green!15}{4.2} & 90.8\,/\,\colorbox{green!26}{1.8} & \colorbox{green!29}{48.5}\,/\,\colorbox{green!14}{10.3} & 26.5\,/\,19.2 & 58.3\,/\,10.1 & \cellcolor{green!25}$+2.2_{\scriptscriptstyle\pm 1.5}$ & \cellcolor{green!23}$-13.1_{\scriptscriptstyle\pm 3.0}$ \\
\cline{2-11}
 & \multirow{2}{*}{AWQ W4} & Base & 43.3\,/\,14.4 & 93.4\,/\,4.0 & 90.6\,/\,2.0 & 47.5\,/\,10.6 & 36.2\,/\,15.8 & 62.2\,/\,9.3 & & \\
 & & +Pen & \colorbox{green!20}{46.7}\,/\,\colorbox{green!13}{13.0} & 93.8\,/\,3.9 & 91.2\,/\,\colorbox{green!10}{1.9} & \colorbox{green!19}{50.5}\,/\,\colorbox{green!11}{9.9} & 34.7\,/\,\colorbox{green!10}{15.2} & 63.4\,/\,8.8 & \cellcolor{green!18}$+1.2_{\scriptscriptstyle\pm 1.7}$ & \cellcolor{green!14}$-5.0_{\scriptscriptstyle\pm 2.9}$ \\
\cline{2-11}
 & \multirow{2}{*}{AWQ W3} & Base & 30.0\,/\,16.1 & 91.2\,/\,4.8 & 90.4\,/\,2.1 & 48.0\,/\,10.7 & 26.9\,/\,21.3 & 57.3\,/\,11.0 & & \\
 & & +Pen & 28.9\,/\,\colorbox{green!13}{14.6} & 90.2\,/\,\colorbox{green!18}{4.0} & 90.5\,/\,\colorbox{green!17}{1.8} & 49.0\,/\,\colorbox{green!11}{10.1} & 28.7\,/\,\colorbox{green!14}{18.9} & 57.5\,/\,9.9 & $+0.2_{\scriptscriptstyle\pm 1.4}$ & \cellcolor{green!22}$-12.0_{\scriptscriptstyle\pm 3.9}$ \\
\cline{2-11}
 & \multirow{2}{*}{FQ W8A8} & Base & 40.0\,/\,13.4 & 93.0\,/\,3.8 & 90.4\,/\,1.9 & 49.5\,/\,9.9 & 35.4\,/\,15.5 & 61.7\,/\,8.9 & & \\
 & & +Pen & \colorbox{green!33}{47.8}\,/\,\colorbox{green!12}{12.3} & 93.8\,/\,\colorbox{green!12}{3.6} & 91.2\,/\,\colorbox{green!18}{1.6} & \colorbox{green!20}{53.0}\,/\,9.9 & \colorbox{green!18}{38.1}\,/\,\colorbox{green!11}{14.6} & 64.8\,/\,8.4 & \cellcolor{green!31}$+3.1_{\scriptscriptstyle\pm 2.4}$ & \cellcolor{green!17}$-7.8_{\scriptscriptstyle\pm 2.5}$ \\
\cline{2-11}
 & \multirow{2}{*}{FQ W4A4} & Base & 23.3\,/\,20.4 & 83.6\,/\,5.4 & 87.7\,/\,2.1 & 50.0\,/\,18.0 & 12.3\,/\,23.9 & 51.4\,/\,14.0 & & \\
 & & +Pen & 22.2\,/\,\colorbox{green!10}{19.7} & \colorbox{green!18}{86.4}\,/\,\colorbox{green!15}{4.7} & 88.5\,/\,\colorbox{green!9}{2.1} & 51.5\,/\,\colorbox{green!18}{14.9} & 14.2\,/\,\colorbox{green!9}{23.2} & 52.6\,/\,12.9 & \cellcolor{green!18}$+1.2_{\scriptscriptstyle\pm 2.4}$ & \cellcolor{green!17}$-7.7_{\scriptscriptstyle\pm 4.8}$ \\
\bottomrule
\end{tabular}%
}
\end{table}

\begin{table}[h]
\centering
\caption{\textbf{Effect of the overthinking penalty on Qwen-14B.} For each quantization configuration, the top row (Base) shows accuracy (\%) and CoT length (thousands of tokens) without penalty, and the bottom row (+Pen) shows results with the best penalty $\lambda$. Per-benchmark cells are colored when the penalty changes accuracy or CoT length by $\geq$2\%. $\Delta$Acc and $\Delta$Len\% report the mean change across the five benchmarks with std across $\lambda$ values.}
\label{app:tab:qwen14b}
\resizebox{\textwidth}{!}{%
\setlength{\tabcolsep}{3.5pt}
\renewcommand{\arraystretch}{1.0}
\scriptsize
\begin{tabular}{cl@{\hskip 4pt}l ccccc c cc}
\toprule
 & &  & \textbf{AIME} & \textbf{MATH} & \textbf{GSM8K} & \textbf{GPQA} & \textbf{LCB} & \textbf{Avg.} & \textbf{$\Delta$Acc} & \textbf{$\Delta$Len\%} \\
\cmidrule(lr){10-10} \cmidrule(lr){11-11}
 & &  & \multicolumn{5}{c}{\footnotesize\textit{Acc (\%)\;/\;Len (k)}} & \footnotesize\textit{Acc\;/\;Len} & \multicolumn{2}{c}{\footnotesize\textit{mean$_{\pm\text{std}}$}} \\
\midrule
\multirow{14}{*}{\rotatebox{90}{Qwen-14B}} & \multirow{2}{*}{\textbf{BF16}} & Base & \textbf{57.8}\,/\,\textbf{13.1} & \textbf{94.6}\,/\,\textbf{3.7} & \textbf{94.1}\,/\,\textbf{1.7} & \textbf{57.6}\,/\,\textbf{9.5} & \textbf{49.6}\,/\,\textbf{14.4} & \textbf{70.7}\,/\,\textbf{8.5} & & \\
 & & +Pen & \colorbox{green!26}{63.3}\,/\,\colorbox{green!19}{10.7} & 95.6\,/\,\colorbox{green!11}{3.5} & 94.5\,/\,\colorbox{green!10}{1.7} & \colorbox{green!25}{62.6}\,/\,\colorbox{green!12}{8.8} & \colorbox{green!24}{54.5}\,/\,\colorbox{green!12}{13.4} & 74.1\,/\,7.6 & \cellcolor{green!33}$+3.4_{\scriptscriptstyle\pm 1.7}$ & \cellcolor{green!18}$-8.5_{\scriptscriptstyle\pm 2.3}$ \\
\cline{2-11}
 & \multirow{2}{*}{GPTQ W4} & Base & 62.2\,/\,11.2 & 95.6\,/\,3.4 & 93.3\,/\,1.6 & 60.1\,/\,9.4 & 52.6\,/\,15.3 & 72.8\,/\,8.2 & & \\
 & & +Pen & \colorbox{red!26}{56.7}\,/\,11.3 & 95.8\,/\,\colorbox{green!13}{3.1} & 94.2\,/\,\colorbox{green!14}{1.5} & \colorbox{green!25}{65.2}\,/\,9.5 & 51.1\,/\,\colorbox{green!11}{14.5} & 72.6\,/\,8.0 & $-0.2_{\scriptscriptstyle\pm 1.8}$ & \cellcolor{green!13}$-4.7_{\scriptscriptstyle\pm 2.7}$ \\
\cline{2-11}
 & \multirow{2}{*}{GPTQ W3} & Base & 44.4\,/\,12.4 & 94.4\,/\,3.6 & 92.8\,/\,1.6 & 53.5\,/\,9.1 & 42.2\,/\,14.6 & 65.5\,/\,8.3 & & \\
 & & +Pen & \colorbox{green!20}{47.8}\,/\,\colorbox{green!10}{11.9} & 93.8\,/\,\colorbox{green!11}{3.4} & 92.5\,/\,\colorbox{green!13}{1.5} & \colorbox{green!17}{56.1}\,/\,\colorbox{green!11}{8.7} & \colorbox{green!25}{47.4}\,/\,14.6 & 67.5\,/\,8.0 & \cellcolor{green!24}$+2.0_{\scriptscriptstyle\pm 1.7}$ & \cellcolor{green!13}$-4.8_{\scriptscriptstyle\pm 2.3}$ \\
\cline{2-11}
 & \multirow{2}{*}{AWQ W4} & Base & 56.7\,/\,12.4 & 95.0\,/\,3.6 & 92.6\,/\,1.6 & 59.6\,/\,9.3 & 48.1\,/\,14.8 & 70.4\,/\,8.4 & & \\
 & & +Pen & 56.7\,/\,\colorbox{green!10}{11.8} & 96.0\,/\,\colorbox{green!13}{3.3} & 94.2\,/\,\colorbox{green!14}{1.5} & \colorbox{green!19}{62.6}\,/\,9.5 & \colorbox{green!22}{52.2}\,/\,\colorbox{green!12}{13.7} & 72.3\,/\,8.0 & \cellcolor{green!23}$+1.9_{\scriptscriptstyle\pm 1.6}$ & \cellcolor{green!15}$-5.9_{\scriptscriptstyle\pm 2.7}$ \\
\cline{2-11}
 & \multirow{2}{*}{AWQ W3} & Base & 48.9\,/\,14.3 & 93.0\,/\,4.3 & 93.0\,/\,2.1 & 56.6\,/\,10.5 & 42.2\,/\,17.5 & 66.7\,/\,9.7 & & \\
 & & +Pen & 47.8\,/\,14.1 & 94.2\,/\,\colorbox{green!13}{3.9} & 92.9\,/\,\colorbox{green!18}{1.7} & 56.1\,/\,\colorbox{green!9}{10.2} & \colorbox{green!21}{45.9}\,/\,\colorbox{green!12}{16.2} & 67.4\,/\,9.2 & \cellcolor{green!14}$+0.6_{\scriptscriptstyle\pm 1.7}$ & \cellcolor{green!16}$-7.5_{\scriptscriptstyle\pm 3.5}$ \\
\cline{2-11}
 & \multirow{2}{*}{FQ W8A8} & Base & 61.1\,/\,11.9 & 94.8\,/\,3.8 & 94.2\,/\,1.7 & 57.6\,/\,9.4 & 51.9\,/\,14.0 & 71.9\,/\,8.2 & & \\
 & & +Pen & 62.2\,/\,\colorbox{green!11}{11.2} & 95.8\,/\,\colorbox{green!12}{3.6} & 94.2\,/\,\colorbox{green!13}{1.6} & \colorbox{green!30}{64.6}\,/\,\colorbox{green!12}{8.8} & 51.9\,/\,\colorbox{green!12}{13.0} & 73.7\,/\,7.6 & \cellcolor{green!22}$+1.8_{\scriptscriptstyle\pm 2.1}$ & \cellcolor{green!16}$-7.1_{\scriptscriptstyle\pm 2.9}$ \\
\cline{2-11}
 & \multirow{2}{*}{FQ W4A4} & Base & 52.2\,/\,12.7 & 95.0\,/\,3.7 & 93.5\,/\,1.8 & 63.6\,/\,8.8 & 49.3\,/\,15.2 & 70.7\,/\,8.4 & & \\
 & & +Pen & 53.3\,/\,12.6 & 94.6\,/\,\colorbox{green!9}{3.6} & 93.9\,/\,\colorbox{green!15}{1.6} & \colorbox{red!16}{61.6}\,/\,8.8 & 49.3\,/\,\colorbox{green!11}{14.3} & 70.5\,/\,8.2 & $-0.2_{\scriptscriptstyle\pm 1.3}$ & \cellcolor{green!13}$-4.5_{\scriptscriptstyle\pm 3.5}$ \\
\bottomrule
\end{tabular}%
}
\end{table}

\begin{table}[h]
\centering
\caption{\textbf{Effect of the overthinking penalty on Llama-8B.} For each quantization configuration, the top row (Base) shows accuracy (\%) and CoT length (thousands of tokens) without penalty, and the bottom row (+Pen) shows results with the best penalty $\lambda$. Per-benchmark cells are colored when the penalty changes accuracy or CoT length by $\geq$2\%. $\Delta$Acc and $\Delta$Len\% report the mean change across the five benchmarks with std across $\lambda$ values.}
\label{app:tab:llama8b}
\resizebox{\textwidth}{!}{%
\setlength{\tabcolsep}{3.5pt}
\renewcommand{\arraystretch}{1.0}
\scriptsize
\begin{tabular}{cl@{\hskip 4pt}l ccccc c cc}
\toprule
 & &  & \textbf{AIME} & \textbf{MATH} & \textbf{GSM8K} & \textbf{GPQA} & \textbf{LCB} & \textbf{Avg.} & \textbf{$\Delta$Acc} & \textbf{$\Delta$Len\%} \\
\cmidrule(lr){10-10} \cmidrule(lr){11-11}
 & &  & \multicolumn{5}{c}{\footnotesize\textit{Acc (\%)\;/\;Len (k)}} & \footnotesize\textit{Acc\;/\;Len} & \multicolumn{2}{c}{\footnotesize\textit{mean$_{\pm\text{std}}$}} \\
\midrule
\multirow{14}{*}{\rotatebox{90}{Llama-8B}} & \multirow{2}{*}{\textbf{BF16}} & Base & \textbf{44.4}\,/\,\textbf{14.5} & \textbf{89.0}\,/\,\textbf{4.6} & \textbf{89.0}\,/\,\textbf{2.2} & \textbf{49.0}\,/\,\textbf{10.2} & \textbf{35.1}\,/\,\textbf{16.2} & \textbf{61.3}\,/\,\textbf{9.6} & & \\
 & & +Pen & \colorbox{red!16}{42.2}\,/\,\colorbox{green!11}{13.8} & \colorbox{green!16}{91.0}\,/\,\colorbox{green!13}{4.2} & 88.6\,/\,\colorbox{green!12}{2.0} & 50.5\,/\,10.1 & \colorbox{green!19}{38.4}\,/\,\colorbox{green!11}{15.3} & 62.1\,/\,9.1 & \cellcolor{green!15}$+0.8_{\scriptscriptstyle\pm 2.0}$ & \cellcolor{green!14}$-5.6_{\scriptscriptstyle\pm 2.7}$ \\
\cline{2-11}
 & \multirow{2}{*}{GPTQ W4} & Base & 41.1\,/\,14.1 & 89.6\,/\,4.6 & 88.6\,/\,2.3 & 48.0\,/\,10.7 & 35.1\,/\,15.5 & 60.5\,/\,9.5 & & \\
 & & +Pen & \colorbox{red!23}{36.7}\,/\,\colorbox{green!9}{13.8} & 89.2\,/\,\colorbox{green!12}{4.3} & 88.6\,/\,\colorbox{green!13}{2.1} & \colorbox{green!22}{52.0}\,/\,\colorbox{green!11}{10.2} & 35.8\,/\,15.8 & 60.5\,/\,9.2 & $+0.0_{\scriptscriptstyle\pm 2.2}$ & \cellcolor{green!13}$-4.7_{\scriptscriptstyle\pm 2.1}$ \\
\cline{2-11}
 & \multirow{2}{*}{GPTQ W3} & Base & 26.7\,/\,18.2 & 82.2\,/\,6.2 & 85.4\,/\,2.4 & 29.8\,/\,12.8 & 24.6\,/\,24.1 & 49.7\,/\,12.7 & & \\
 & & +Pen & \colorbox{green!16}{28.9}\,/\,\colorbox{green!14}{16.2} & 83.4\,/\,\colorbox{green!13}{5.6} & 85.8\,/\,\colorbox{green!15}{2.0} & \colorbox{green!40}{39.9}\,/\,12.7 & \colorbox{green!16}{26.9}\,/\,\colorbox{green!10}{23.0} & 53.0\,/\,11.9 & \cellcolor{green!32}$+3.2_{\scriptscriptstyle\pm 1.9}$ & \cellcolor{green!17}$-7.9_{\scriptscriptstyle\pm 3.9}$ \\
\cline{2-11}
 & \multirow{2}{*}{AWQ W4} & Base & 30.0\,/\,15.5 & 89.2\,/\,4.8 & 88.0\,/\,2.2 & 48.0\,/\,10.5 & 35.1\,/\,16.9 & 58.1\,/\,10.0 & & \\
 & & +Pen & \colorbox{green!16}{32.2}\,/\,\colorbox{green!16}{13.2} & \colorbox{red!19}{86.2}\,/\,\colorbox{green!12}{4.4} & 88.1\,/\,\colorbox{green!17}{1.8} & \colorbox{green!17}{50.5}\,/\,10.4 & 35.4\,/\,\colorbox{green!12}{15.7} & 58.5\,/\,9.1 & \cellcolor{green!13}$+0.4_{\scriptscriptstyle\pm 1.6}$ & \cellcolor{green!19}$-9.3_{\scriptscriptstyle\pm 2.5}$ \\
\cline{2-11}
 & \multirow{2}{*}{AWQ W3} & Base & 12.2\,/\,14.8 & 80.4\,/\,5.4 & 82.6\,/\,2.4 & 35.9\,/\,9.6 & 23.1\,/\,18.1 & 46.8\,/\,10.1 & & \\
 & & +Pen & \colorbox{green!23}{16.7}\,/\,\colorbox{green!12}{13.6} & \colorbox{red!18}{77.6}\,/\,\colorbox{green!15}{4.7} & 83.7\,/\,\colorbox{green!18}{2.0} & 36.9\,/\,\colorbox{green!14}{8.6} & 23.1\,/\,\colorbox{green!15}{15.9} & 47.6\,/\,9.0 & \cellcolor{green!15}$+0.7_{\scriptscriptstyle\pm 1.5}$ & \cellcolor{green!22}$-12.2_{\scriptscriptstyle\pm 2.7}$ \\
\cline{2-11}
 & \multirow{2}{*}{FQ W8A8} & Base & 37.8\,/\,14.4 & 88.8\,/\,4.6 & 89.6\,/\,2.1 & 48.5\,/\,10.6 & 37.3\,/\,15.6 & 60.4\,/\,9.5 & & \\
 & & +Pen & \colorbox{green!16}{40.0}\,/\,\colorbox{green!10}{13.9} & 89.6\,/\,\colorbox{green!11}{4.3} & 89.2\,/\,\colorbox{green!11}{2.0} & 50.0\,/\,\colorbox{green!12}{9.8} & \colorbox{green!17}{39.9}\,/\,15.7 & 61.7\,/\,9.1 & \cellcolor{green!19}$+1.3_{\scriptscriptstyle\pm 1.7}$ & \cellcolor{green!13}$-4.7_{\scriptscriptstyle\pm 2.6}$ \\
\cline{2-11}
 & \multirow{2}{*}{FQ W4A4} & Base & 22.2\,/\,15.4 & 86.4\,/\,5.0 & 87.6\,/\,2.4 & 40.9\,/\,10.2 & 35.1\,/\,17.8 & 54.4\,/\,10.2 & & \\
 & & +Pen & 22.2\,/\,\colorbox{green!10}{14.7} & \colorbox{red!16}{84.2}\,/\,4.9 & 86.1\,/\,\colorbox{green!12}{2.2} & \colorbox{green!32}{48.5}\,/\,\colorbox{green!10}{9.8} & \colorbox{red!20}{31.7}\,/\,\colorbox{green!9}{17.3} & 54.5\,/\,9.8 & $+0.1_{\scriptscriptstyle\pm 1.7}$ & \cellcolor{green!12}$-4.1_{\scriptscriptstyle\pm 3.0}$ \\
\bottomrule
\end{tabular}%
}
\end{table}

\begin{table}[h]
\centering
\caption{\textbf{Effect of the overthinking penalty on QwQ-32B.} For each quantization configuration, the top row (Base) shows accuracy (\%) and CoT length (thousands of tokens) without penalty, and the bottom row (+Pen) shows results with the best penalty $\lambda$. Per-benchmark cells are colored when the penalty changes accuracy or CoT length by $\geq$2\%. $\Delta$Acc and $\Delta$Len\% report the mean change across the five benchmarks with std across $\lambda$ values.}
\label{app:tab:qwq32b}
\resizebox{\textwidth}{!}{%
\setlength{\tabcolsep}{3.5pt}
\renewcommand{\arraystretch}{1.0}
\scriptsize
\begin{tabular}{cl@{\hskip 4pt}l ccccc c cc}
\toprule
 & &  & \textbf{AIME} & \textbf{MATH} & \textbf{GSM8K} & \textbf{GPQA} & \textbf{LCB} & \textbf{Avg.} & \textbf{$\Delta$Acc} & \textbf{$\Delta$Len\%} \\
\cmidrule(lr){10-10} \cmidrule(lr){11-11}
 & &  & \multicolumn{5}{c}{\footnotesize\textit{Acc (\%)\;/\;Len (k)}} & \footnotesize\textit{Acc\;/\;Len} & \multicolumn{2}{c}{\footnotesize\textit{mean$_{\pm\text{std}}$}} \\
\midrule
\multirow{10}{*}{\rotatebox{90}{QwQ-32B}} & \multirow{2}{*}{\textbf{BF16}} & Base & \textbf{76.7}\,/\,\textbf{15.3} & \textbf{97.2}\,/\,\textbf{4.5} & \textbf{95.7}\,/\,\textbf{1.9} & \textbf{63.1}\,/\,\textbf{10.1} & \textbf{62.3}\,/\,\textbf{17.9} & \textbf{79.0}\,/\,\textbf{9.9} & & \\
 & & +Pen & \colorbox{green!23}{81.1}\,/\,\colorbox{green!19}{12.3} & 98.0\,/\,\colorbox{green!14}{4.0} & 96.0\,/\,\colorbox{green!12}{1.8} & \colorbox{green!20}{66.7}\,/\,\colorbox{green!11}{9.5} & 61.9\,/\,\colorbox{green!10}{17.3} & 80.7\,/\,9.0 & \cellcolor{green!22}$+1.7_{\scriptscriptstyle\pm 1.2}$ & \cellcolor{green!19}$-9.6_{\scriptscriptstyle\pm 2.6}$ \\
\cline{2-11}
 & \multirow{2}{*}{GPTQ W4} & Base & 78.9\,/\,15.7 & 97.4\,/\,4.5 & 95.6\,/\,1.9 & 59.6\,/\,9.9 & 58.6\,/\,19.0 & 78.0\,/\,10.2 & & \\
 & & +Pen & \colorbox{red!16}{76.7}\,/\,\colorbox{green!14}{13.9} & 97.8\,/\,\colorbox{green!10}{4.2} & 96.1\,/\,\colorbox{green!13}{1.7} & \colorbox{green!23}{64.1}\,/\,\colorbox{green!11}{9.4} & \colorbox{green!17}{61.2}\,/\,\colorbox{green!12}{17.7} & 79.2\,/\,9.4 & \cellcolor{green!18}$+1.2_{\scriptscriptstyle\pm 1.1}$ & \cellcolor{green!17}$-7.7_{\scriptscriptstyle\pm 2.7}$ \\
\cline{2-11}
 & \multirow{2}{*}{GPTQ W3} & Base & 64.4\,/\,16.1 & 96.8\,/\,4.9 & 95.4\,/\,2.0 & 56.1\,/\,9.3 & 50.4\,/\,19.1 & 72.6\,/\,10.3 & & \\
 & & +Pen & 65.6\,/\,\colorbox{green!12}{15.0} & 96.4\,/\,\colorbox{green!14}{4.3} & 95.9\,/\,\colorbox{green!12}{1.8} & \colorbox{green!22}{60.1}\,/\,\colorbox{red!9}{9.5} & 51.9\,/\,\colorbox{green!11}{18.0} & 74.0\,/\,9.7 & \cellcolor{green!19}$+1.3_{\scriptscriptstyle\pm 1.4}$ & \cellcolor{green!15}$-5.9_{\scriptscriptstyle\pm 3.9}$ \\
\cline{2-11}
 & \multirow{2}{*}{AWQ W4} & Base & 75.6\,/\,15.7 & 97.4\,/\,4.5 & 95.8\,/\,1.9 & 62.6\,/\,10.8 & 59.0\,/\,19.1 & 78.1\,/\,10.4 & & \\
 & & +Pen & \colorbox{green!23}{80.0}\,/\,\colorbox{green!13}{14.2} & 96.8\,/\,\colorbox{green!12}{4.2} & 95.9\,/\,\colorbox{green!12}{1.8} & \colorbox{green!19}{65.7}\,/\,\colorbox{red!9}{11.0} & 60.8\,/\,18.8 & 79.8\,/\,10.0 & \cellcolor{green!22}$+1.8_{\scriptscriptstyle\pm 1.1}$ & \cellcolor{green!13}$-4.8_{\scriptscriptstyle\pm 2.8}$ \\
\cline{2-11}
 & \multirow{2}{*}{AWQ W3} & Base & 60.0\,/\,16.7 & 96.0\,/\,4.6 & 95.9\,/\,1.8 & 61.1\,/\,10.1 & 51.5\,/\,19.5 & 72.9\,/\,10.5 & & \\
 & & +Pen & \colorbox{green!16}{62.2}\,/\,\colorbox{green!15}{14.5} & 96.6\,/\,\colorbox{green!13}{4.2} & 95.5\,/\,\colorbox{green!12}{1.7} & 60.6\,/\,\colorbox{green!16}{8.7} & 53.4\,/\,\colorbox{green!10}{18.7} & 73.7\,/\,9.5 & \cellcolor{green!15}$+0.7_{\scriptscriptstyle\pm 1.2}$ & \cellcolor{green!19}$-9.6_{\scriptscriptstyle\pm 2.9}$ \\
\bottomrule
\end{tabular}%
}
\end{table}

\section{Token Lists Used in Experiments}
\label{app:thinking-tokens}

We report the four token lists (50 tokens each) used in our experiments. The overthinking markers list is used for the main penalty intervention. The other three lists are used as controls in the ablation study (\Cref{sec:ablation}). Tokens prefixed with ``\_'' include a leading space in the tokenizer, making them distinct token IDs from their space-free variants.

\paragraph{Overthinking markers (50 tokens).} Lexical items that semantically express hesitation, redirection, or local backtracking. The list is constructed based on our inspection of tokens frequently sampled at high-KL decoding positions and is consistent with prior work that defines a similar class of hesitation tokens \citep{zhao2025let}.

\begin{table}[h!]
\centering
\small
\setlength{\tabcolsep}{6pt}
\begin{tabular}{p{0.94\linewidth}}
\toprule
\textbf{Overthinking markers (50 tokens)} \\
\midrule
\texttt{\_perhaps}, \texttt{\_maybe}, \texttt{\_wait}, \texttt{\_Wait}, \texttt{\_actually}, \texttt{\_hold}, \texttt{\_Hmm}, \texttt{\_hmm}, \texttt{\_Alternatively}, \texttt{\_alternatively}, \texttt{\_However}, \texttt{\_however}, \texttt{\_instead}, \texttt{\_Instead}, \texttt{\_But}, \texttt{\_but}, \texttt{\_though}, \texttt{\_although}, \texttt{\_yet}, \texttt{\_rather}, \texttt{\_unless}, \texttt{\_otherwise}, \texttt{\_nonetheless}, \texttt{\_nevertheless}, \texttt{\_regardless}, \texttt{\_still}, \texttt{\_anyway}, \texttt{\_Or}, \texttt{\_or}, \texttt{\_either}, \texttt{\_whether}, \texttt{\_uncertain}, \texttt{\_unsure}, \texttt{\_possibly}, \texttt{\_might}, \texttt{\_could}, \texttt{\_another}, \texttt{\_different}, \texttt{\_reconsider}, \texttt{\_rethink}, \texttt{\_backtrack}, \texttt{\_retry}, \texttt{\_recheck}, \texttt{\_revisit}, \texttt{\_doubt}, \texttt{\_confused}, \texttt{\_wrong}, \texttt{\_mistake}, \texttt{\_error}, \texttt{\_incorrect}. \\
\bottomrule
\end{tabular}
\caption{Overthinking markers used in the penalty intervention. All 50 tokens include a leading space (denoted by ``\_'').}
\label{app:tab:thinking-tokens}
\end{table}

\paragraph{High-KL tokens (50 tokens).} The 50 tokens with the highest average KL divergence between BF16 and AWQ 3-bit on MATH-500, filtered to tokens appearing at least 50 times. This list overlaps partially with the overthinking markers but also includes functional tokens such as ``\_So'' and ``\_The'' that are unrelated to overthinking.

\begin{table}[h!]
\centering
\small
\setlength{\tabcolsep}{6pt}
\begin{tabular}{p{0.94\linewidth}}
\toprule
\textbf{High-KL tokens (50 tokens)} \\
\midrule
\texttt{\_using}, \texttt{\_if}, \texttt{\_Since}, \texttt{\_But}, \texttt{\_That}, \texttt{\_Wait}, \texttt{\_since}, \texttt{\_If}, \texttt{\_seems}, \texttt{\_maybe}, \texttt{\_not}, \texttt{Alternatively}, \texttt{But}, \texttt{\_The}, \texttt{\_more}, \texttt{Wait}, \texttt{So}, \texttt{\_verify}, \texttt{\_let}, \texttt{\_in}, \texttt{\_So}, \texttt{First}, \texttt{\_problem}, \texttt{\_each}, \texttt{\_another}, \texttt{\_but}, \texttt{\_\textbackslash n\textbackslash n}, \texttt{\_Then}, \texttt{\_think}, \texttt{\_it}, \texttt{\_all}, \texttt{\_Let}, \texttt{\_at}, \texttt{\_I}, \texttt{\_there}, \texttt{\_sum}, \texttt{\_compute}, \texttt{\_other}, \texttt{\_gives}, \texttt{\_Therefore}, \texttt{\_then}, \texttt{\_make}, \texttt{\_must}, \texttt{\_see}, \texttt{\_correct}, \texttt{\_how}, \texttt{\_the}, \texttt{\_adding}, \texttt{\_because}, \texttt{\_have}. \\
\bottomrule
\end{tabular}
\caption{The 50 tokens with the highest average KL divergence between BF16 and AWQ 3-bit. Tokens with ``\_'' have a leading space.}
\label{app:tab:highkl-tokens}
\end{table}

\paragraph{Low-KL tokens (50 tokens).} The 50 tokens with the lowest average KL divergence, filtered to tokens appearing at least 50 times. These are predominantly mathematical symbols, digits, and formatting tokens.

\begin{table}[h!]
\centering
\small
\setlength{\tabcolsep}{6pt}
\begin{tabular}{p{0.94\linewidth}}
\toprule
\textbf{Low-KL tokens (50 tokens)} \\
\midrule
\texttt{\_Answer}, \texttt{</think>}, \texttt{<EOS>}, \texttt{[\textbackslash n}, \texttt{\{}, \texttt{\}\textbackslash}, \texttt{aps}, \texttt{Final}, \texttt{**\textbackslash n}, \texttt{sqrt}, \texttt{boxed}, \texttt{b}, \texttt{\{\textbackslash}, \texttt{\}\{}, \texttt{\_than}, \texttt{times}, \texttt{]\textbackslash n\textbackslash n}, \texttt{k}, \texttt{5}, \texttt{frac}, \texttt{(x}, \texttt{text}, \texttt{6}, \texttt{x}, \texttt{\_\_}, \texttt{3}, \texttt{8}, \texttt{\}}, \texttt{2}, \texttt{)\^{}}, \texttt{9}, \texttt{0}, \texttt{1}, \texttt{a}, \texttt{\_<}, \texttt{\_sure}, \texttt{'t}, \texttt{)}, \texttt{\_power}, \texttt{4}, \texttt{\}\textbackslash n}, \texttt{\textbackslash [}, \texttt{\_\textbackslash}, \texttt{\_-}, \texttt{\_+}, \texttt{\_b}, \texttt{\^{}}, \texttt{\textbackslash}, \texttt{(\textbackslash}. \\
\bottomrule
\end{tabular}
\caption{The 50 tokens with the lowest average KL divergence. These mathematical and formatting tokens remain stable under quantization.}
\label{app:tab:lowkl-tokens}
\end{table}

\paragraph{Random tokens (50 tokens).} 50 randomly selected tokens serving as a null control.

\begin{table}[h!]
\centering
\small
\setlength{\tabcolsep}{6pt}
\begin{tabular}{p{0.94\linewidth}}
\toprule
\textbf{Random tokens (50 tokens)} \\
\midrule
\texttt{\_Course}, \texttt{\_Examples}, \texttt{\_Entity}, \texttt{\_teacher}, \texttt{\_");}, \texttt{\_-type}, \texttt{\_James}, \texttt{\_softly}, \texttt{\_";}, \texttt{\_yat}, \texttt{\_:The}, \texttt{\_Mend}, \texttt{\_Proposed}, \texttt{\_.ident}, \texttt{\_elda}, \texttt{\_(\_;,}, \texttt{\_preseason}, \texttt{\_ID}, \texttt{\_sort}, \texttt{\_fn}, \texttt{\_eng}, \texttt{\_Emoji}, \texttt{\_accessing}, \texttt{\_urniture}, \texttt{\_Normally}, \texttt{\_REUTERS}, \texttt{\_notifying}, \texttt{\_ResourceManager}, \texttt{\_delayed}, \texttt{\_Applying}, \texttt{\_supporter}, \texttt{\_Hot}, \texttt{\_build}, \texttt{\_PERTIES}, \texttt{\_noticing}, \texttt{\_crops}, \texttt{\_goalie}, \texttt{\_Petr}, \texttt{\_\_commit}, \texttt{\_.validators}, \texttt{\_.Trace}, \texttt{\_cut}, \texttt{\_propose}, \texttt{\_Porno}, \texttt{\_point}, \texttt{\_zend}, \texttt{\_Call}, \texttt{\_(pass}, \texttt{\_UserId}, \texttt{\_picture}. \\
\bottomrule
\end{tabular}
\caption{50 randomly selected tokens used as a null control. All 50 include a leading space (``\_'').}
\label{app:tab:random-tokens}
\end{table}

\section{GPT-5 Prompt for Error Categorization}
\label{app:prompt-error-categorization}

We provide the prompt used to categorize incorrect generations into the four error types reported in \Cref{fig:error_breakdown} and \Cref{fig:error_gsm8k}. The prompt instantiates three fields: (i) the original problem, (ii) the model-generated reasoning trace and final answer, and (iii) the gold answer. The judge is instructed to assign \emph{exactly one} label among the four categories and to select \emph{overthinking} whenever the gold answer appears in the intermediate reasoning but is not produced as the final answer. Otherwise, the model has to categorize the error in the other 3 categories. Among all experiments conducted using GPT-5 as judge, the model never failed to choose exactly one category and return it. The judge achieves >95\% agreement with human annotations on a subset of the incorrect questions.

\begin{table}[h!]
\centering
\small
\setlength{\tabcolsep}{6pt}
\begin{tabular}{p{0.94\linewidth}}
\toprule
\textbf{Prompt for GPT-5 Judge} \\
\midrule
You are an expert at analyzing mathematical reasoning traces from a given model to identify error types. \\

\vspace{0.1em}

Classify the error into exactly ONE of these categories: \\

\vspace{0.1em}

(A) overthinking: The model reaches the correct solution at some point but does NOT commit to it. It instead opens new reasoning paths or excessively questions the assumptions that it made. \\
\vspace{0.1em}
(B) logical\_error: The model follows an incorrect plan or misunderstands the problem from early on. \\
\vspace{0.1em}
(C) arithmetic\_error: The overall approach/plan is correct, but the model makes concrete computational/calculation mistakes. \\
\vspace{0.1em}
(D) formatting\_hallucination\_other: The model produces an off-format answer, introduces hallucinated constraints, or exhibits other behavioral failures not covered above. \\

\vspace{0.1em}
Problem: \\
{problem} \\

\vspace{0.1em}
Model's Reasoning and Final Answer: \\
{model\_output} \\

\vspace{0.1em}
Correct (Gold) Answer: \\
{gold\_answer} \\

\vspace{0.1em}
If at any point in the chain-of-thought, the model finds the correct (gold) answer but does not commit to it as the final answer, choose overthinking (A). Otherwise, analyze the reasoning and determine if the error is a logical error (B), an arithmetic error (C), or a formatting/hallucination/other error (D). Respond with your final answer as a single letter: A, B, C, or D. \\

\bottomrule
\end{tabular}
\caption{Prompt template used for GPT-5 error categorization.}
\label{fig:prompt-error-categorization}
\end{table}

\end{document}